\documentclass{article}

\usepackage{graphicx}

 \usepackage[eandd,final,nonatbib]{neurips_2026}
\usepackage[numbers,square]{natbib}
\usepackage[utf8]{inputenc} 
\usepackage[T1]{fontenc}    
\usepackage{hyperref}       
\usepackage{url}            
\usepackage{booktabs}       
\usepackage{multirow}
\usepackage{amsfonts}       
\usepackage{amsmath}
\usepackage{nicefrac}       
\usepackage{microtype}      
\usepackage{xcolor}         
\usepackage{enumitem}
\usepackage{listings}
\usepackage{pgfplots}
\pgfplotsset{compat=1.17}
\usepgfplotslibrary{groupplots}

\newtheorem{definition}{Definition}

\title{Federated Dataset Simulation: Inducing Label-Free Heterogeneity Across Tasks}

\author{%
  Vasilis Siomos$^{1,2}$\thanks{Corresponding author.}\hspace{1em}Lam Ngo$^{1,3,4}$\hspace{1em}Jonathan Passerat-Palmbach$^{1,5,6}$\hspace{1em}Giacomo Tarroni$^{1,6}$ \\
  $^{1}$Department of Computer Science, City St George's, University of London \quad $^{2}$Ultromics \\
  $^{3}$ELLIS Institute Finland \\
  $^{4}$Department of Electrical Engineering and Automation, Aalto University \quad $^{5}$Flashbots \\
  $^{6}$Department of Computing, Imperial College London \\
  \texttt{vasilis.siomos@ultromics.com} \quad \texttt{lam.ngo@aalto.fi} \\
  \texttt{j.passerat-palmbach@imperial.ac.uk} \quad \texttt{giacomo.tarroni@citystgeorges.ac.uk}
}

\begin{document}

\maketitle

\begin{abstract}
Federated Learning enables collaborative model training across decentralized clients without sharing raw data. While the field has seen rapid growth, the datasets underpinning FL evaluation have not kept pace: real federated datasets remain scarce, and most studies rely on small-scale centralized classification datasets partitioned via Dirichlet label skew, a one-dimensional proxy for the multi-dimensional heterogeneity present in real federations. Moreover, label-based partitioning is by construction inapplicable to non-classification tasks such as segmentation and detection, for which no principled simulation method exists. We cast the conversion of centralized datasets into federated benchmarks as the task of Federated Dataset Simulation (FDS), and decompose it into a representation function $\Phi$ and an assignment mechanism $\pi$. We provide two label-free instantiations---geometric partitioning via pretrained model embeddings and semantic partitioning via vision-language model-derived criteria---applicable to classification and dense prediction tasks alike. We evaluate across four datasets spanning classification, segmentation, and detection, and show that both methods induce heterogeneity that matches or exceeds that of label-skew partitioning, with measurable impact on downstream FL performance. We release \texttt{fds}, a modular open-source package for end-to-end federated dataset simulation, heterogeneity measurement, and FL training, at \url{https://github.com/siomvas/fds}.
\end{abstract}
\section{Introduction}\label{sec:intro}
Federated learning (FL) \cite{fedavg} is a collaborative training paradigm that has matured into a central solution for privacy-preserving and decentralized ML, with production deployments in several industries \cite{fl_explosion}. Despite an exponential growth in research output, a well-documented gap persists in the scale, complexity, and diversity of the datasets and tasks underpinning FL evaluation \cite{flower_datasets, survey_simple}: the vast majority of empirical studies are conducted on image classification at the scale and complexity of CIFAR \cite{cifar}. While naturally federated datasets exist \cite{leaf,flamby,fets}, expanding this dataset pool is laborious and costly, and the heterogeneity such datasets exhibit is often idiosyncratic and not transferable to other tasks or domains. It has therefore become standard practice to partition existing centralized datasets into virtual clients for FL evaluation \cite{jimenez_survey}. For a virtual partition to provide a challenging FL evaluation, the consensus is it must reflect a central tenet of FL theory, namely that federated optimization performance degrades as the statistical heterogeneity (non-IIDness) of client data increases \cite{fedavg,fieldguide}. 

This creates a need for methods simulating non-IIDness across the virtual clients of a split centralized dataset. The prevailing approach to do so is Dirichlet-based label partitioning \cite{jimenez_survey}, which assigns skewed class proportions across clients by sampling from a Dirichlet distribution \cite{hsu2019measuring}. This reduces the multi-dimensional heterogeneity (covariate shift, concept shift, domain shift) observed in real federations to a single axis, when in practice clients may differ not only in which classes they observe but in how those classes appear \cite{kairouz}. Moreover, Dirichlet partitioning operates on sample-level class labels, so is by construction inapplicable to any task where samples are not uniquely associated with a single category (e.g., segmentation, machine translation, and sound event detection). For these tasks, which constitute a growing share of practical FL applications, no principled heterogeneity simulation method exists, and practitioners are limited to random partitioning or ad-hoc splits \cite{survey_simple}. As a result, current FL evaluations and benchmarks produce conclusions about FL research progress (in performance, convergence rates, fairness, and robustness) that may not generalize to real deployments.

We cast this problem as the task of \textbf{Federated Dataset Simulation (FDS)}: converting a centralized dataset into a federated benchmark in a controllable manner, making heterogeneity simulation itself an object of systematic design and measurement. To enable simulation that is agnostic to the downstream task and label format, we decompose FDS into two sub-tasks: a representation function $\Phi$, which maps data into a structured latent space that captures axes of variation related to heterogeneity, and an assignment mechanism $\pi$, which allocates samples to virtual clients within that space. This $\Phi/\pi$ decomposition provides a unifying formalism under which existing approaches, like label Dirichlet partitioning, can be expressed as special cases. Because $\Phi$ operates on raw inputs rather than annotations, the framework is naturally label-free and modality-agnostic. We provide and evaluate two concrete instantiations of this framework in the vision domain, each targeting a different latent space. The first, \textit{geometric partitioning}, defines $\Phi_{\text{geo}}$ as the embedding function of a self-supervised pretrained foundation model and uses $k$-means clustering as the assignment $\pi$. Because such encoders are trained to organize semantically related inputs into coherent regions of their representation space, partitioning along geometric boundaries in this space induces covariate shift that reflects meaningful visual and semantic structure. The second, \textit{semantic partitioning}, defines $\Phi_{\text{sem}}$ as a chain of pretrained vision-language and large language models which extracts per-sample sets of human-interpretable semantic criterion labels. A Dirichlet-weighted assignment $\pi$ then correlates client identity with specific criterion values to produce federations where each client can be characterized by the visual properties they tend to exhibit. 

We validate the framework across four datasets encompassing classification, semantic segmentation, and object detection, and benchmark FL performance under a varying number of clients. We demonstrate that both geometric and semantic partitioning produce distinct heterogeneity profiles that surpass existing approaches, not only in terms of feature and label (where applicable) heterogeneity, but also on FL convergence. Our contributions are as follows: 
\begin{enumerate}[leftmargin=*,itemsep=0pt, topsep=0pt]
    \item \textbf{Task formulation}. We formalize federated dataset simulation through the $\Phi/\pi$ decomposition, establishing a unified framework in which heterogeneity type, severity, and measurement are jointly specified and become the object of study and design.
    \item \textbf{Partitioning methods}. We introduce two distinct label-free partitioning pipelines that induce feature-level heterogeneity and apply to classification and dense prediction tasks alike, resulting in similar or greater heterogeneity than Dirichlet partitioning despite using no labels.
    \item \textbf{Empirical evaluation and benchmark}. We produce a multi-faceted evaluation spanning label and feature space heterogeneity along with FL performance and validate the framework across four challenging vision datasets, showing that the proposed partitioning methods result in a more challenging setting that can serve as a better benchmark for FL algorithm development.
    \item \textbf{Software package}. With the intention of lowering the barrier for researchers to move beyond label-skew-only evaluation and classification, we release \texttt{fds}, a modular open-source library providing an end-to-end pipeline for federated dataset simulation: feature extraction across multiple pretrained encoders, semantic clustering based on our $\Phi_{\text{sem}}$ VLM-LLM pipeline, controllable partitioning to clients, heterogeneity measurement in the feature or label space, and integrated downstream FL training for classification, segmentation, or detection.
\end{enumerate}
\section{The FDS Framework}\label{sec:framework}
\subsection{General Problem Formulation}
\begin{definition}[Federated Dataset Simulation]
Let $\mathcal{D} = \{(x_i, y_i)\}_{i=1}^N$ denote a centralized dataset comprising $N$ samples, where $x_i \in \mathcal{X}$ and $y_i \in \mathcal{Y}$ represent the input and annotation space, respectively. The Federated Dataset Simulation (FDS) task seeks a partition $\{\mathcal{D}_k\}_{k=1}^K$ such that: $\bigcup_k \mathcal{D}_k = \mathcal{D}$ (all samples are assigned), $\mathcal{D}_k \cap \mathcal{D}_{\ell \neq k} = \emptyset$  (samples are uniquely assigned), and the induced local distributions $P_k(x,y)$ exhibit meaningful statistical heterogeneity.
\end{definition}

\begin{definition}[Partitioning Mechanism]
Given $\mathcal{D}$, a partitioning mechanism consists of:
\begin{enumerate}[leftmargin=*]
    \setlength\itemsep{0em}
    \item A representation function $\Phi: \mathcal{X \times \mathcal{Y}} \to \mathcal{Z}$, which projects inputs into a structured latent space $\mathcal{Z}$.
    \item An assignment mechanism $\pi$ which produces a partitioning $\{\mathcal{D}_k\}_{k=1}^K$ from the set of latent representations $\{z_i=\Phi(x_i)\}_{i=1}^N$.
\end{enumerate}
\end{definition}

The assignment mechanism operates via a set of client parameters $\omega = \{\omega_k\}_{k=1}^{K}$, which may be computed from the data, sampled from a prior, or both, and then allocates samples to clients based on the affinity between each sample's representation and the client parameters. Label-skew Dirichlet partitioning \cite{hsu2019measuring} fits this framework as the special case where $\Phi(x)=y$, $\omega_k=\mathbf{q}_k\sim\text{Dir}(\alpha\mathbf{p})$ and samples of each class are allocated to clients in proportion to $\{q_{k,y}\}_{k=1}^{K}$. The concentration parameter $\alpha$ controls heterogeneity from IID ($\alpha \to \infty$) to near class-disjoint ($\alpha \to 0$). 

Unlike the Dirichlet special case, the two instantiations we propose next operate on $\Phi(x)$ alone, making them applicable to settings where sample-level labels are unavailable, noisy, or dense (e.g., segmentation and detection tasks). This also extends the space of heterogeneity that can be expressed: because the label space $\mathcal{Y}$ is a lossy projection of $\mathcal{X}$, feature-based representations can capture distributional structure that label-based partitioning cannot.

\begin{definition}[Heterogeneity Functional]
Given a realized partition, we quantify heterogeneity via the mean pairwise divergence of client distributions in the representation space:

$$\mathcal{H}(\pi, \Phi) = \frac{1}{\binom{K}{2}} \sum_{k < l} D_\Phi(P_k, P_\ell)$$

\end{definition}
where $P_k$ denotes the empirical distribution of $\Phi(x)$ for samples assigned to client $k$, and $D_\Phi$ is a divergence measure appropriate to $\mathcal{Z}$.

\textbf{Label Distribution Heterogeneity} When sample-level class labels are available, we examine how client label distributions diverge. We measure this via the Jensen--Shannon divergence, a symmetric, bounded measure well-suited to discrete distributions: 

$$\mathrm{JSD}(P_k^y \| P_l^y)={\frac {1}{2}}D_{\text{KL}}(P_k^y\| M)+{\frac {1}{2}}D_{\text{KL}}(P_l^y\parallel M), \quad M=(P_k^y+P_l^y) \;/ \;2$$

\textbf{Feature-distribution heterogeneity} To measure heterogeneity in a continuous representation space, we use the Maximum Mean Discrepancy (MMD) \cite{gretton2012kernel} with an RBF kernel and the Frechet Inception Distance (FID) \cite{fid}, computed in the activation space of a frozen InceptionV3 network~\cite{szegedy2016rethinking} independent of the embedding encoder used for partitioning, avoiding the circularity of evaluating partitions in the same embedding space used to create them: 

\begin{align*}
  \mathrm{MMD}^2 &= \| \mu_{P_k} - \mu_{P_l} \|_{\mathcal{F}}^2, \text{ where } P\mu_P \text{ is the kernel mean embedding of } P   \\
  \mathrm{FID} &= \| \mu_k - \mu_l \|^2 + \mathrm{Tr}(\Sigma_k + \Sigma_l - 2(\Sigma_k \Sigma_l)^{1/2}
\end{align*}

\subsection{Geometric Partitioning} 
This instantiation operates in the continuous feature space induced by pretrained foundation models. Let $f_\theta: \mathcal{X} \to \mathbb{R}^d$ be a model pretrained with a self-supervised or vision-language objective. We define $\Phi_{\text{geo}}(x)$ as $f_\theta(x) \in \mathbb{R}^d$. The assignment mechanism partitions the embedding space via $k$-means clustering with $K$ clusters equal to the number of target clients. The client parameters are the cluster centroids $\omega = \{\mu_k\}_{k=1}^{K}$, computed from the full set of representations, and each sample is assigned to its nearest centroid $k_i = \arg\min_{k \in [K]} \|\Phi_{\text{geo}}(x_i) - \mu_k\|_2$. 

Modern pretrained representations, whether trained via self-distillation \cite{dinov2} or vision-language alignment \cite{CLIP}, organize semantically related inputs into coherent regions of the embedding space, so partitioning along geometric boundaries in this space induces covariate shift that reflects the visual and semantic structure captured by the encoder. The resulting clients specialize in different visual domains, object types, or scene compositions, depending on which axes of variation are salient in the encoder's representation space, without requiring access to annotations. The number of clients $K$ acts as an implicit heterogeneity knob, where fewer clients produce broader clusters spanning more diverse manifold regions, while more clients yield finer partitions with greater inter-client divergence. We evaluate multiple encoder families and token aggregation strategies in Section \ref{sec:eval} and report the selection procedure used to identify a recommended default configuration.

\subsection{VLM-based Semantic Partitioning}
The second instantiation operates over a discrete criterion space $\mathcal{Z}_{\text{sem}} = \mathcal{L}_1 \times \mathcal{L}_2 \times \cdots \times \mathcal{L}_C$, where each $\mathcal{L}_c$ is a finite set of categorical values corresponding to a human-interpretable semantic criterion (e.g., cuisine origin or texture for a food photography dataset). The representation function $\Phi_{\text{sem}}: \mathcal{X} \to \mathcal{Z}_{\text{sem}}$ maps each image to a tuple of criterion labels $\Phi_{\text{sem}}(x_i) = (\ell_1(x_i), \ldots, \ell_C(x_i))$, and the assignment mechanism $\pi$ allocates samples to clients based on these discrete representations using a product-of-Dirichlets mechanism.

\textbf{Criterion label extraction} $\Phi_\text{sem}$ is a multi-stage pipeline chaining a vision-language model (VLM) and a large language model (LLM) to perform text-conditioned image clustering. For each dataset, semantically meaningful axes of variation are identified through domain inspection, and candidate criteria are validated by querying the VLM on a sample subset to verify non-degenerate label distributions. After $C$ is established, per-sample labeling proceeds as an iterative procedure, inspired by \citet{ictc}: the VLM is prompted to produce a short description of each image focusing exclusively on criterion $c$. An LLM converts each description into a single-word raw label summarizing the image with respect to $c$ to form a dictionary. The LLM then clusters this dictionary of raw labels into $|\mathcal{L}_c|$ canonical categories to consolidate semantically close concepts. Each sample description is then classified into one of these $|\mathcal{L}_c|$ categories, yielding the criterion label $\ell_c(x_i) \in \mathcal{L}_c$, using the same LLM. This is repeated for each criterion to construct $\Phi_{\text{sem}}(x_i)$.

\textbf{Assignment mechanism} Given the extracted criterion vectors, samples are assigned to clients via a product-of-Dirichlets scheme. For each client $m$ and each criterion $c$, a preference vector $\boldsymbol{p}_{m,c} \sim \mathrm{Dir}(\alpha \cdot \mathbf{1}_{|\mathcal{L}_c|})$ is drawn. The assignment probability of sample $i$ to client $m$ is $\pi_m \propto  \prod_{c=1}^{C} p_{m,c}\bigl[\ell_c(x_i)\bigr]$, where $p_{m,c}[\ell]$ denotes the preference weight that client $m$ places on value $\ell$ of criterion $c$. This multiplicative composition ensures that a sample receives high affinity for a client only if it matches that client's preferences across all criteria simultaneously. This mechanism generalizes the standard Dirichlet label-skew assignment to the multi-label setting. The concentration $\alpha$ serves as the heterogeneity knob, like in the typical label-skew case. The full set of criteria and prompt details used per dataset are reported in Appendix \ref{app:vlm} and included in our code release. As this is a computationally expensive pipeline, our code release also includes the precomputed criterion labels for all 4 evaluated datasets.

\subsection{Connection to Federated Learning Performance}\label{subsec:con_to_fl}
The previous sections define methods for inducing distributional divergence across virtual clients, but not all such divergence is equally consequential for federated optimization. A partition that separates clients along axes to which the model can easily learn invariant features may produce large $\mathcal{H}$ while having negligible impact on model convergence. The optimization difficulty induced by a partition depends not only on the magnitude of $\mathcal{H}$ but on the mutual information $I(\mathcal{Z}; Y)$ between the representation and the task: when $I(\mathcal{Z}; Y) \approx 0$ heterogeneity in $\mathcal{Z}$ does not translate into optimization difficulty as opposed to when $I(\mathcal{Z}; Y)$ is high. This clarifies why ad-hoc heterogeneity methods such as additive Gaussian noise \cite{li2022federated} or geometric rotations \cite{ghosh2020efficient} do not typically produce meaningful FL difficulty especially if the training pipeline includes augmentations for such perturbations. To effectively benchmark FL algorithms across different tasks, client configurations, and optimization strategies, we therefore need partitioning methods that produce heterogeneity with high $I(\mathcal{Z}; Y)$.

Both proposed instantiations are designed to meet this requirement. $\Phi_{\text{geo}}$ is extracted from encoders whose pretraining objectives aim to preserve semantic content, producing representations with high $I(\mathcal{Z};Y)$ across a wide range of vision tasks \cite{ssl_dark_matter}. Partitioning in this space induces distributional shifts that the model is likely to be sensitive to, which we verify empirically in Section \ref{sec:eval}. For $\Phi_{\text{sem}}$, task-coupling depends on the choice of criteria, with criteria that describe properties correlated with the target variable (e.g., main object type, scene composition) producing higher $I(\mathcal{Z}_{\text{sem}}; Y)$ than criteria more orthogonal to the task, as we show via criterion-level ablation in Section \ref{sec:eval}. This reflects an asymmetry between the two instantiations: geometric partitioning derives task-relevance implicitly from the pretraining objective, while semantic partitioning requires the explicit design of task-relevant criteria but offers interpretability, as the axes of heterogeneity are human-readable.
\section{Evaluation}\label{sec:eval}


\begin{table*}[t]
\centering
\caption{Heterogeneity metrics on classification datasets across client counts $K \in \{5, 10, 20\}$. JSD measures label-distribution divergence; MMD, FID, and centroid distance measure feature-distribution divergence (computed in InceptionV3 activation space). Higher values indicate greater heterogeneity. Geometric partitioning uses no labels yet induces the highest heterogeneity across all metrics, including JSD.}
\label{tab:het_cls}
\small
\setlength{\tabcolsep}{3pt}
\begin{tabular}{ll rrr rrr rrr rrr}
\toprule
& & \multicolumn{3}{c}{JSD $\uparrow$} & \multicolumn{3}{c}{MMD ($\times 10^{-2}$) $\uparrow$} & \multicolumn{3}{c}{FID $\uparrow$} & \multicolumn{3}{c}{Centroid $\uparrow$} \\
\cmidrule(lr){3-5} \cmidrule(lr){6-8} \cmidrule(lr){9-11} \cmidrule(lr){12-14}
Dataset & Method & \small 5 & \small 10 & \small 20 & \small 5 & \small 10 & \small 20 & \small 5 & \small 10 & \small 20 & \small 5 & \small 10 & \small 20 \\
\midrule
\multirow{5}{*}{Tiny}
& Random               & .003 & .005 & .010 & 0.00 & 0.00 & 0.00 &   3.3 &   6.6 &  13.2 &  0.9 &  0.9 &  0.9 \\
& Label Dir.\ $\alpha$=0.1   & .559 & .572 & .569 & 0.40 & 0.55 & 0.75 &  29.2 &  39.0 &  49.4 &  4.9 &  6.0 &  7.0 \\
& Semantic $\alpha$=1.0 & .128 & .168 & .207 & 0.70 & 0.93 & 1.19 &  22.0 &  30.9 &  43.0 &  4.8 &  5.5 &  6.2 \\
& Semantic $\alpha$=0.1 & .152 & .240 & .315 & 0.58 & 1.33 & 2.15 &  23.2 &  42.1 &  62.2 &  4.8 &  6.7 &  8.7 \\
& Geometric            & \textbf{.630} & \textbf{.655} & \textbf{.668} & \textbf{5.31} & \textbf{9.70} & \textbf{18.0} & \textbf{111} & \textbf{146} & \textbf{201} & \textbf{13.8} & \textbf{21.6} & \textbf{31.1} \\
\midrule
\multirow{5}{*}{Food}
& Random               & .002 & .003 & .007 & 0.00 & 0.00 & 0.00 &   2.7 &   5.4 &  10.5 &  0.8 &  0.9 &  1.0 \\
& Label Dir.\ $\alpha$=0.1   & .564 & .587 & .566 & 0.69 & 1.06 & 1.06 &  16.5 &  24.6 &  29.9 &  8.3 &  9.9 & 11.1 \\
& Semantic $\alpha$=1.0 & .138 & .168 & .215 & 0.67 & 0.75 & 1.10 &  12.5 &  16.3 &  25.7 &  8.0 &  8.2 &  9.7 \\
& Semantic $\alpha$=0.1 & .213 & .284 & .359 & 0.61 & 1.50 & 2.47 &  14.3 &  25.3 &  39.8 &  8.3 & 11.2 & 14.3 \\
& Geometric            & \textbf{.620} & \textbf{.660} & \textbf{.664} & \textbf{6.58} & \textbf{16.4} & \textbf{13.8} & \textbf{74.0} & \textbf{127} & \textbf{122} & \textbf{27.0} & \textbf{39.0} & \textbf{40.5} \\
\bottomrule
\end{tabular}
\end{table*}


\begin{table*}[t]
\centering
\caption{Heterogeneity metrics on dense prediction datasets. Dirichlet label-skew partitioning is inapplicable to segmentation and detection, so Random is the only existing baseline. Geometric and semantic partitioning are the first methods to produce non-trivial heterogeneity for these tasks.}
\label{tab:het_dense}
\small
\setlength{\tabcolsep}{4pt}
\begin{tabular}{ll rrr rrr rrr}
\toprule
& & \multicolumn{3}{c}{MMD ($\times 10^{-2}$) $\uparrow$} & \multicolumn{3}{c}{FID $\uparrow$} & \multicolumn{3}{c}{Centroid $\uparrow$} \\
\cmidrule(lr){3-5} \cmidrule(lr){6-8} \cmidrule(lr){9-11}
Dataset & Method & \small 5 & \small 10 & \small 20 & \small 5 & \small 10 & \small 20 & \small 5 & \small 10 & \small 20 \\
\midrule
\multirow{4}{*}{ADE20K}
& Random               & 0.00 & 0.00 & 0.01 &  11.1 &  21.8 &  41.3 &  0.9 &  1.3 &  1.8 \\
& Semantic $\alpha$=1.0 & 2.13 & 2.27 & 2.16 &  47.1 &  60.8 &  80.5 &  6.4 &  6.8 &  7.0 \\
& Semantic $\alpha$=0.1 & 2.80 & 4.56 & 5.10 &  59.4 &  92.6 & 124.6 &  8.1 & 10.1 & 11.0 \\
& Geometric            & \textbf{13.1} & \textbf{16.5} & \textbf{19.0} & \textbf{167} & \textbf{191} & \textbf{247} & \textbf{20.7} & \textbf{25.6} & \textbf{31.9} \\
\midrule
\multirow{4}{*}{Pascal VOC}
& Random               & 0.01 & 0.02 & 0.02 &  52.3 &  87.4 & 129.6 &  2.0 &  2.8 &  3.9 \\
& Semantic $\alpha$=1.0 & 0.68 & 1.21 & 1.29 &  68.6 & 112.2 & 152.1 &  5.1 &  6.7 &  7.6 \\
& Semantic $\alpha$=0.1 & 1.09 & 1.52 & 2.16 &  79.8 & 120.5 & 170.6 &  6.3 &  7.5 &  9.4 \\
& Geometric            & \textbf{18.1} & \textbf{16.7} & \textbf{30.1} & \textbf{219} & \textbf{228} & \textbf{267} & \textbf{22.0} & \textbf{23.6} & \textbf{29.7} \\
\bottomrule
\end{tabular}
\end{table*}

%
%
%
\subsection{Evaluation Design}

\textbf{Encoder and model selection.}
Our geometric partitioning method depends on the choice of representation function $\Phi_{\mathrm{geo}}$. We therefore evaluate several pretrained vision encoders, including DINOv2~\cite{dinov2}, CLIP\cite{CLIP}, and MAE\cite{MAE} variants across several token extraction and aggregation strategies, and select a single default configuration based on average rank of MMD across datasets. Unless otherwise stated, we use DINOv2-B features obtained by concatenating the CLS tokens from the final four layers, followed by UMAP \cite{mcinnes2018umap} dimensionality reduction before clustering. This choice provided a strong and computationally tractable default across classification, segmentation, and detection tasks. For semantic partitioning, we use the open-source Intern-VL3.5~\cite{internvl3} and Llama 3.3 70B~\cite{llama} models as the VLM and LLM components, respectively, in order to keep the pipeline reproducible and accessible under a modest compute budget; all extracted criterion labels are cached and reused across experiments.

\textbf{Datasets and tasks.}
We evaluate on four vision datasets spanning three prediction settings. Tiny-ImageNet~\cite{le2015tiny} provides a more visually complex classification benchmark than CIFAR-scale datasets while remaining comparable to prior FL work. Food101~\cite{bossard2014food} introduces a fine-grained, domain-specific classification task with substantial intra-class variation and label noise. To test whether FDS extends beyond sample-level class labels, we further include ADE20K~\cite{ade20k} for semantic segmentation and PASCAL VOC\cite{everingham2010pascal} for object detection. These dense prediction tasks are important for our setting because standard label Dirichlet partitioning is not naturally defined when each sample contains multiple labels, spatially structured labels, or object sets rather than a single class.

\textbf{FL training protocol.}
For each dataset, we compare random partitioning, label Dirichlet partitioning where applicable, geometric partitioning, and semantic partitioning. We vary the number of clients $K \in \{5,10,20\}$ to study how each partitioning mechanism scales with federation size. All FL experiments use FedAvg~\citep{fedavg} with 1 local epoch for 100 communication rounds. We use a ResNet-18\cite{resnet}, DeepLabv3 with ResNet-50~\cite{chen2017rethinking} for semantic segmentation, and a Faster R-CNN model with a ResNet-50-FPN backbone \cite{li2021benchmarking} for detection. All models are initialized from pretrained weights. We tune dataset-specific optimization hyperparameters on a held-out subset using Optuna~\cite{akiba2019optuna} and then fix them across partitioning methods.

\textbf{Metrics.}
We evaluate each partition along three axes. We measure feature-space heterogeneity using the MMD, FID, and feature centroid Euclidean distance averaged over pairwise distances between client distributions in the encoder representation space. Where sample-level labels are available, we report label-space heterogeneity using Jensen--Shannon divergence between client label distributions. Downstream FL performance is evaluated using task-appropriate metrics: top-1 accuracy for Tiny ImageNet and Food101, mIoU for ADE20K, and mAP@50 for PASCAL VOC. This separates the question of whether a partition induces measurable distributional heterogeneity from whether that heterogeneity produces optimization difficulty for FL.

\begin{table*}[t]
\centering
\caption{
Downstream FL performance under FedAvg across datasets and client counts. Parentheses report degradation relative to random partitioning. Lower non-random performance within each dataset/client-count block in bold. Label Dirichlet baselines are omitted for dense prediction tasks because they do not provide a single sample-level class label.
}
\label{tab:fl_all}
\small
\setlength{\tabcolsep}{5pt}
\begin{tabular}{lllccc}
\toprule
Dataset & Metric & Partition & $K=5$ & $K=10$ & $K=20$ \\
\midrule
\multirow{6}{*}{Tiny ImageNet}
& \multirow{6}{*}{Acc.}
& Random
& 74.53 (0.00) & 74.63 (0.00) & 74.29 (0.00) \\
&& Label Dir. $\alpha=0.1$
& 71.74 (2.79) & 71.56 (3.07) & 71.74 (2.55) \\
&& Label Dir. $\alpha=1.0$
& 74.30 (0.23) & 74.19 (0.44) & 73.57 (0.72) \\
&& Semantic $\alpha=0.1$
& 73.69 (0.84) & 70.43 (4.20) & 72.99 (1.30) \\
&& Semantic $\alpha=1.0$
& 73.56 (0.97) & 71.32 (3.31) & 72.31 (1.98) \\
&& Geometric
& \textbf{67.41 (7.12)} & \textbf{64.84 (9.79)} & \textbf{69.24 (5.05)} \\
\midrule
\multirow{6}{*}{Food101}
& \multirow{6}{*}{Acc.}
& Random
& 77.16 (0.00) & 74.43 (0.00) & 70.39 (0.00) \\
&& Label Dir. $\alpha=0.1$
& 71.31 (5.85) & 66.45 (7.98) & 64.20 (6.19) \\
&& Label Dir. $\alpha=1.0$
& 75.75 (1.41) & 72.89 (1.54) & 68.80 (1.59) \\
&& Semantic $\alpha=0.1$
& 72.93 (4.23) & 64.49 (9.94) & \textbf{63.97 (6.42)} \\
&& Semantic $\alpha=1.0$
& 74.93 (2.23) & 71.66 (2.77) & 65.68 (4.71) \\
&& Geometric
& \textbf{70.02 (7.14)} & \textbf{63.15 (11.28)} & 66.34 (4.05) \\
\midrule
\multirow{4}{*}{ADE20K}
& \multirow{4}{*}{mIoU}
& Random
& 40.23 (0.00) & 39.02 (0.00) & 36.91 (0.00) \\
&& Semantic $\alpha=0.1$
& 38.13 (2.10) & 36.74 (2.28) & 33.46 (3.45) \\
&& Semantic $\alpha=1.0$
& 38.98 (1.25) & 37.82 (1.20) & 34.01 (2.90) \\
&& Geometric
& \textbf{35.37 (4.86)} & \textbf{31.14 (7.88)} & \textbf{29.34 (7.57)} \\
\midrule
\multirow{4}{*}{PASCAL VOC}
& \multirow{4}{*}{mAP@50}
& Random
& 79.40 (0.00) & 83.10 (0.00) & 82.14 (0.00) \\
&& Semantic $\alpha=0.1$
& 78.15 (1.25) & 77.56 (5.54) & 77.99 (4.15) \\
&& Semantic $\alpha=1.0$
& 78.12 (1.28) & 78.12 (4.98) & 79.22 (2.92) \\
&& Geometric
& \textbf{74.34 (5.06)} & \textbf{73.19 (9.91)} & \textbf{78.10 (4.04)} \\
\bottomrule
\end{tabular}
\end{table*}
\subsection{Main Results}

\textbf{Induced Heterogeneity} Tables~\ref{tab:het_cls} and~\ref{tab:het_dense} show that the proposed FDS instantiations induce feature-space heterogeneity across tasks. On classification datasets, label Dirichlet partitioning strongly increases label-distribution divergence but has comparatively modest effect in feature space. In contrast, geometric partitioning, despite using no labels, produces the largest feature-space divergence and also induces high label divergence as a byproduct of partitioning semantically structured representations. Semantic partitioning occupies an intermediate regime: it produces non-random feature heterogeneity with an interpretable severity parameter, and the $\alpha=0.1$ setting consistently increases heterogeneity relative to $\alpha=1.0$. On dense prediction datasets both geometric and semantic partitioning produce non-trivial feature heterogeneity relative to random splits, supporting the use of the $\Phi/\pi$ framework beyond classification.

\textbf{Impact on downstream FL performance.}
The heterogeneity induced by FDS translates into measurable downstream optimization difficulty. Table~\ref{tab:fl_all} reports global validation performance under FedAvg for $K \in \{5,10,20\}$ clients. Across all datasets and client counts, non-random partitions degrade FedAvg performance relative to random splitting, indicating that the induced client differences are not merely artifacts of the heterogeneity metrics. Geometric partitioning is the most challenging setting in most configurations, producing the largest degradation on all Tiny ImageNet, ADE20K, and PASCAL VOC settings, and on Food101 for $K=5$ and $K=10$. Semantic partitioning is generally less severe but remains consistently harder than random, while preserving an interpretable $\alpha$-controlled assignment mechanism. On classification datasets, label-Dirichlet skew remains a strong baseline at $\alpha=0.1$; however, on dense prediction tasks it is not directly applicable, whereas both proposed FDS instantiations continue to induce non-trivial performance degradation.

\begin{table}[t]
\centering
\caption{Accuracy (\%) of seven aggregation methods on Tiny ImageNet with $K = 10$, from random and from ImageNet initializations. Settings differ from Table~\ref{tab:fl_all} (Appendix~\ref{app:hparam}). Mean $\pm$ standard deviation over three seeds, rank in parentheses. Geometric partitioning is the most challenging protocol for every method.}
\label{tab:algos}
\setlength{\tabcolsep}{3pt}
\resizebox{\linewidth}{!}{%
\begin{tabular}{@{}lcccccc@{}}
\toprule
 & \multicolumn{3}{c}{Random initialization} & \multicolumn{3}{c}{Pretrained weights} \\
\cmidrule(lr){2-4}\cmidrule(lr){5-7}
Method & Label Dir. & Geometric & Semantic & Label Dir. & Geometric & Semantic \\
\midrule
FedAvg & 32.97{\scriptsize$\pm$0.09} (3) & 24.89{\scriptsize$\pm$0.22} (4) & 35.14{\scriptsize$\pm$0.35} (2) & 54.73{\scriptsize$\pm$0.23} (3) & 42.11{\scriptsize$\pm$0.16} (4) & 55.97{\scriptsize$\pm$0.34} (5) \\
FedDyn & 30.92{\scriptsize$\pm$0.13} (6) & 25.11{\scriptsize$\pm$0.04} (2) & 33.91{\scriptsize$\pm$0.08} (5) & 53.06{\scriptsize$\pm$0.10} (7) & 36.41{\scriptsize$\pm$0.33} (6) & 56.73{\scriptsize$\pm$0.02} (3) \\
FedNova & 32.91{\scriptsize$\pm$0.36} (4) & 20.03{\scriptsize$\pm$7.69} (7) & 33.76{\scriptsize$\pm$0.06} (6) & 54.22{\scriptsize$\pm$0.05} (4) & 26.32{\scriptsize$\pm$1.86} (7) & 55.72{\scriptsize$\pm$0.39} (7) \\
FedProx & 33.07{\scriptsize$\pm$0.09} (2) & 24.92{\scriptsize$\pm$0.24} (3) & 34.98{\scriptsize$\pm$0.09} (4) & 54.17{\scriptsize$\pm$0.12} (5) & 42.95{\scriptsize$\pm$0.08} (2) & 55.98{\scriptsize$\pm$0.08} (4) \\
FedSAM & 30.16{\scriptsize$\pm$0.18} (7) & \textbf{25.97}{\scriptsize$\pm$0.10} (1) & 32.66{\scriptsize$\pm$0.26} (7) & 55.45{\scriptsize$\pm$0.03} (2) & \textbf{45.66}{\scriptsize$\pm$0.08} (1) & 57.27{\scriptsize$\pm$0.01} (2) \\
SCAFFOLD & \textbf{33.16}{\scriptsize$\pm$0.10} (1) & 24.50{\scriptsize$\pm$3.57} (6) & \textbf{35.94}{\scriptsize$\pm$0.15} (1) & \textbf{55.63}{\scriptsize$\pm$0.45} (1) & 40.89{\scriptsize$\pm$2.56} (5) & \textbf{58.37}{\scriptsize$\pm$0.18} (1) \\
FedLC & 32.27{\scriptsize$\pm$0.16} (5) & 24.80{\scriptsize$\pm$0.30} (5) & 35.09{\scriptsize$\pm$0.17} (3) & 53.35{\scriptsize$\pm$0.13} (6) & 42.42{\scriptsize$\pm$0.14} (3) & 55.84{\scriptsize$\pm$0.37} (6) \\
\midrule
Spread & 3.00 & 5.94 & 3.28 & 2.57 & 19.34 & 2.65 \\
\bottomrule
\end{tabular}}
\end{table}

\subsection{Ranking of Aggregation Methods}\label{sec:algos}
We compare seven aggregation methods, FedAvg~\citep{fedavg}, FedDyn~\citep{feddyn}, FedNova~\citep{fednova}, FedProx~\citep{fedprox}, FedSAM~\citep{fedsam}, SCAFFOLD~\citep{scaffold}, and FedLC~\citep{fedlc}, under label Dirichlet skew and semantic partitioning at $\alpha = 0.1$ and under geometric partitioning, on Tiny ImageNet with 10 clients and a ResNet-18 trained from random initialization or ImageNet weights (Table~\ref{tab:algos}, settings in Appendix~\ref{app:hparam}). Geometric partitioning is the hardest protocol for every method, and it also orders the methods differently.

Semantic partitioning preserves much of the ranking that label skew produces, with Spearman correlations of 0.750 from random initialization and 0.464 from pretrained weights, whereas the ranking under geometric partitioning correlates with it at $-0.571$ and $0.179$. FedSAM, last of the seven under label skew from random initialization, is first under geometric partitioning from either starting point, while FedNova falls to last. From pretrained weights the seven lie within 2.57 points under label skew, each within a point of its nearest neighbor, whereas under geometric partitioning they spread over 19.34 points, or 9.25 without FedNova, and SCAFFOLD, which edges FedSAM by 0.18 points under label skew, trails it by 4.77. From random initialization the spread widens from 3.00 to 5.94 points only because of FedNova, without which it narrows to 1.47.

\begin{table}[t]
\centering
\caption{
Natural versus simulated heterogeneity. For each naturally federated dataset, we compare the original client split to a geometric partition with the same number of clients. Higher values indicate greater heterogeneity.
}
\label{tab:natural-comparison}
\resizebox{0.8\linewidth}{!}{
\begin{tabular}{llccc}
\toprule
Dataset & Partition & Label JSD & Feature MMD & $k$-NN consistency \\
\midrule
\multirow{2}{*}{Fed-ISIC2019}
& Natural & \textbf{0.115} & 0.027 & 0.260 \\
& Geometric & 0.028 & \textbf{0.192} & \textbf{0.530} \\
\midrule
\multirow{2}{*}{Office-Caltech10}
& Natural & 0.010 & 0.016 & 0.446 \\
& Geometric & \textbf{0.684} & \textbf{0.334} &\textbf{ 0.982} \\
\midrule
\multirow{2}{*}{FEMNIST}
& Natural & 0.130 & 0.043 & 0.127 \\
& Geometric & \textbf{0.643} & \textbf{0.465} & \textbf{0.654} \\
\bottomrule
\end{tabular}}
\end{table}
\subsection{Comparison to Naturally Federated Datasets}\label{sec:natural}
We next ask how the magnitude of heterogeneity induced by representation-space partitioning compares against that naturally present in some naturally multi-centric/federated datasets. To do this we compare the feature space and label space heterogeneity induced by $\Phi_{\text{geo}}$ in naturally federated datasets with the heterogeneity originally present, as measured by JSD, MMD, and k-NN consistency, which is the average fraction of nearest neighbors in feature space belonging to the same client, measuring local clustering structure in feature space \cite{knn_consistency}. We choose Fed-ISIC2019, a skin lesion classification dataset with 6 clients~\citep{flamby}, FEMNIST, a handwriting dataset of digits and letters separated by writer with 3,550 users \citep{leaf}, and Office-Caltech10~\citep{gong2012geodesic}, a domain generalization dataset of 4 distinct domains (Amazon, Caltech, DSLR, and Webcam) with one client per domain, and compare their original client partitions against geometric partitioning under $\Phi_{\text{geo}}$ with the same number of clients. Table~\ref{tab:natural-comparison} shows that geometric partitioning often matches or exceeds the natural client split in feature space. On FEMNIST and Office-Caltech10, geometric partitioning produces substantially larger MMD and $k$-NN client consistency than the natural partition. On Fed-ISIC2019, the natural split exhibits higher label-space divergence, while geometric partitioning produces larger feature-space separation. This distinction is important: real federations can be heterogeneous for different reasons, and the proposed framework makes the choice of representation explicit rather than reducing all heterogeneity to label imbalance.

\subsection{Ablating the Assignment Mechanism in Semantic Partitioning}
\label{sec:criterion-ablation}
To validate that the VLM-extracted criterion labels capture meaningful axes of variation and to examine which criteria contribute most to heterogeneity, we conduct a pure-criterion ablation. For each criterion $c$ and dataset, we create $|L_c|$ clients by deterministically assigning all samples sharing the same criterion value to the same client, bypassing the stochastic Dirichlet assignment entirely. This represents the maximum heterogeneity achievable along a single semantic axis and isolates the contribution of each criterion independently of the assignment mechanism. Table~\ref{tab:criterion-ablation} (Appendix~\ref{app:ablation}) reports feature-space heterogeneity under these pure-criterion splits across all four datasets. A consistent hierarchy emerges: criteria describing \emph{what} appears in an image (main object, scene, cuisine) produce the largest distributional divergence, while criteria describing \emph{how} it appears (lighting, color, size) produce substantially less. This ordering is consistent across metrics and aligns with the intuition that content-level properties carry more mutual information with the task label than stylistic ones, as discussed in Section~\ref{subsec:con_to_fl}. To test whether the feature-space hierarchy translates into downstream optimization difficulty, we train FedAvg on Tiny ImageNet under each pure-criterion split. The ranking is preserved exactly: \texttt{main\_object}, the strongest criterion in feature space, yields the lowest accuracy (67.58\%), a degradation comparable to geometric partitioning at $K{=}5$ (67.41\%, Table~\ref{tab:fl_all}), while \texttt{time}, the weakest, degrades accuracy by only 2.47 points to 72.06\%. MMD and accuracy degradation across the five criteria are perfectly correlated, consistent with criteria with higher $I(Z_{\mathrm{sem}}; Y)$ producing more consequential heterogeneity (Section~\ref{subsec:con_to_fl}).
\textbf{Validating the pipeline.}\label{sec:validation} On 10,531 Clevr-4 images~\citep{vaze2023norep}, whose shape, texture, color, and count taxonomies are equally valid, the label of each criterion agrees with its own taxonomy far more than with the others, with NMI from 0.297 for count to 0.860 for shape, against 0.012 on average off the diagonal. Our Food101 cuisine labels reach an AMI of 0.503 with the human labels of Food-4c~\citep{liu2026organizing} (Appendix~\ref{app:validation}).

\begin{figure}[t]
\centering
\begin{tikzpicture}
\begin{groupplot}[group style={group size=2 by 1, horizontal sep=1.3cm}, width=0.4\linewidth, height=2.8cm,
  xmode=normal, xtick={0,1,2,3,5,10}, xlabel={$p$}, ylabel={NMI}, ymin=0, ymax=1,
  title style={font=\small, yshift=-4pt}, tick label style={font=\scriptsize}, label style={font=\small},
  legend style={font=\scriptsize, draw=none}]
\nextgroupplot[title={DINOv2}, legend to name=projlegend, legend columns=1]
\addplot[color=orange!85!black, mark=triangle*, mark size=1.4pt, thick] coordinates {(0,0.0346) (1,0.0273) (2,0.0403) (3,0.0232) (5,0.1942) (10,0.4277)};
\addlegendentry{texture}
\addplot[color=blue!70!black, mark=*, mark size=1.4pt, thick] coordinates {(0,0.9279) (1,0.9195) (2,0.8865) (3,0.8771) (5,0.5171) (10,0.1862)};
\addlegendentry{shape}
\addplot[color=green!50!black, mark=square*, mark size=1.4pt, thick] coordinates {(0,0.0014) (1,0.0019) (2,0.0029) (3,0.0022) (5,0.0060) (10,0.0113)};
\addlegendentry{color}
\addplot[color=gray, mark=diamond*, mark size=1.4pt, thick] coordinates {(0,0.0064) (1,0.0070) (2,0.0049) (3,0.0065) (5,0.0780) (10,0.0150)};
\addlegendentry{count}
\nextgroupplot[title={MAE}]
\addplot[color=orange!85!black, mark=triangle*, mark size=1.4pt, thick] coordinates {(0,0.0209) (1,0.0054) (2,0.0088) (3,0.0143) (5,0.0413) (10,0.0496)};
\addplot[color=blue!70!black, mark=*, mark size=1.4pt, thick] coordinates {(0,0.4085) (1,0.5249) (2,0.5991) (3,0.4149) (5,0.1624) (10,0.1145)};
\addplot[color=green!50!black, mark=square*, mark size=1.4pt, thick] coordinates {(0,0.0733) (1,0.1407) (2,0.1539) (3,0.3090) (5,0.4957) (10,0.3057)};
\addplot[color=gray, mark=diamond*, mark size=1.4pt, thick] coordinates {(0,0.1467) (1,0.0973) (2,0.0071) (3,0.0086) (5,0.0107) (10,0.0246)};
\end{groupplot}
\node[anchor=west] at ([xshift=4pt]group c2r1.east) {\ref*{projlegend}};
\end{tikzpicture}
\caption{NMI between $k$-means partitions ($K=10$) of Clevr-4 and each of its taxonomies after removing the top $p$ principal components of DINOv2 (left) or MAE (right) embeddings. The $k$-means seed standard deviation at $p=0$ is at most 0.021 (DINOv2, three seeds). No partition collapses, since every client holds at least 1.2\% of the images, and the coefficient of variation of client sizes ranges from 0.22 to 0.37 for MAE and from 0.27 to 0.61 for DINOv2.}
\label{fig:projection}
\end{figure}
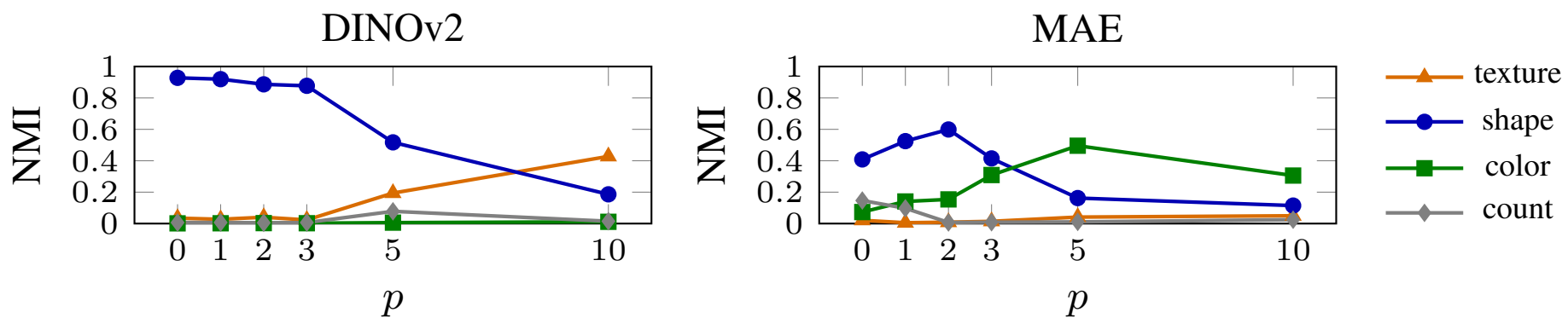
\subsection{What Geometric Partitioning Separates}\label{sec:geometry}
Geometric partitioning requires no criteria, leaving open what its clients differ in. Against the pure-criterion partitions, they follow content far more than appearance, with a mean AMI at $K = 10$ of 0.377 for main object, scene, and cuisine against 0.060 for lighting, color, camera, and size (Appendix~\ref{app:composition}).

The content that a partition follows is set by the representation. Removing the top $p$ principal components before clustering shifts the partition among the properties that the encoder captures (Figure~\ref{fig:projection}). For DINOv2 it raises the NMI with texture from 0.035 at $p = 0$ to 0.428 at $p = 10$ while that with shape falls from 0.928 to 0.186, but leaves color below 0.012, and for MAE it raises the NMI with color from 0.073 to 0.496 at $p = 5$, far beyond the variation across $k$-means seeds. Encoder pretraining thus fixes what a representation can express while the projection fixes what it does express, and $p$ controls the axis of heterogeneity as $\alpha$ controls its severity.

\subsection{Discussion}
\textbf{Feature-space and label-space heterogeneity are not interchangeable.} Label Dirichlet partitioning at $\alpha{=}0.1$ produces high label divergence on classification datasets but comparatively modest feature-space divergence: on Tiny ImageNet with $K{=}10$, it achieves JSD 0.572 but only MMD $0.55 \times 10^{-2}$, whereas geometric partitioning reaches JSD 0.655 alongside MMD $9.70 \times 10^{-2}$. This indicates that label-skew partitions create clients that differ in which classes they hold but not in how they look, limiting the diversity of optimization challenges they present. Evaluation conducted solely under label skew therefore exercises a single axis of heterogeneity, and Section~\ref{sec:algos} shows that heterogeneity along a different axis can change which aggregation method appears best.

\textbf{Not all heterogeneity is equally consequential for federated optimization.} The criterion ablation (Section~\ref{sec:criterion-ablation}) shows that the relationship between distributional divergence and FL difficulty is mediated by the mutual information between the heterogeneity axis and the task, the \texttt{main\_object} criterion on Tiny ImageNet costing 6.95 points of accuracy against the random baseline while \texttt{time} costs 2.47. Effective heterogeneity simulation therefore requires partitioning along task-relevant axes, not merely maximizing distributional distance.

\textbf{Dense prediction tasks are underserved by current simulation practice.} On ADE20K and PASCAL~VOC, where label Dirichlet partitioning is not directly applicable, random splitting has been the only available baseline. Random splits leave these tasks untested under heterogeneity, and geometric partitioning degrades mIoU on ADE20K by up to 7.88 points and mAP@50 on PASCAL~VOC by up to 9.91 points relative to random. Given the growing importance of segmentation and detection in deployed FL systems, this gap represents a blind spot in current evaluation methodology that the $\Phi / \pi$ framework directly addresses.

\textbf{Geometric and semantic partitioning occupy complementary roles.} Geometric partitioning produces the most challenging FL setting in most configurations and requires no design decisions beyond encoder selection, making it a strong default for stress-testing FL algorithms. Semantic partitioning trades peak heterogeneity for interpretability: the $\alpha$ parameter provides continuous control, and the criterion-level ablation allows practitioners to understand and select the axes along which their federation is heterogeneous. The choice between the two depends on whether the goal is maximal difficulty or controlled, explainable variation. Semantic partitioning thus extends the prevailing protocol to tasks it cannot reach, while geometric partitioning creates a regime in which aggregation methods rank differently than under label skew (Appendix~\ref{app:guidance} gives practical guidance).

\section{Related Work}\label{sec:related}
Federated dataset simulation is contemporaneous with FL itself, as the original FedAvg experiments constructed virtual non-IID clients by assigning each client a number of label shards \citep{fedavg}. Subsequent work by \citet{hsu2019measuring} systematized and popularized the Dirichlet label-skew allocation first introduced in \citet{yurochkin2019bayesian}, constructing a tunable protocol where the Dirichlet concentration $\alpha$ acts as a continuous control on client heterogeneity. While this remains the dominant synthetic partitioning protocol in empirical FL \citep{jimenez_survey,survey_simple}, it is limited by construction to representing client heterogeneity through variations in label marginals and therefore assumes single, sample-level class labels. Attempts to simulate feature heterogeneity instead have largely remained at the level of input perturbations, including client-specific Gaussian noise \citep{li2022federated}, image rotations \citep{ghosh2020efficient}, corruption-based perturbations \citep{fang2023}, affine transformations \citep{reisizadeh2020robust}, and Dirichlet allocation over binned feature values \citep{fedartml}. These methods are generally not guaranteed to affect task-relevant representations, making their induced heterogeneity difficult to interpret, measure, or transfer across tasks and label formats.

More recently, \citet{borazjani} explore the use of learned embedding spaces for partitioning data into non-IID clients. Their procedure fine-tunes a task-specific model, extracts latent representations, clusters them with $k$-means, and treats the resulting clusters as pseudo-labels for a standard Dirichlet split. They demonstrate that representation-aware heterogeneity can expose failure modes missed by class-based partitions across multiple aggregation methods. However, the number of pseudo-labels $k$ is hard to interpret, and this method relies on task-specific training, making the resulting partitions difficult to standardize across studies. In contrast, we formulate federated dataset simulation as a decomposition into a representation function $\Phi$ and an assignment mechanism $\pi$, enabling label-free simulation in both continuous and semantic representation spaces and supporting evaluation through feature-space, label-space, and downstream FL metrics.
\section{Conclusion}\label{sec:conclusion}
We introduce Federated Dataset Simulation (FDS), a framework that decomposes heterogeneity simulation into a representation function $\Phi$ and an assignment mechanism $\pi$, unifying existing label-based approaches with novel feature-level and semantic-level partitioning strategies. Through two concrete instantiations and an accompanying evaluation suite, we provide the community with a comprehensive benchmark for controlled heterogeneity research. The $\Phi/\pi$ decomposition is inherently extensible with new representation functions and assignment mechanisms.

\textbf{Broader impact.} FDS lets privacy-preserving FL algorithms be stress-tested under controllable heterogeneity along several dimensions, for dense prediction as well as classification, without access to naturally federated data. The open-source release is intended to lower the barrier for FL researchers to move beyond label-skew-only evaluation.

\textbf{Limitations.} The current evaluation is restricted to computer vision tasks but extending the framework to text, tabular, and multi-modal data is a natural direction for future work. Greater feature-space heterogeneity and greater FedAvg degradation do not by themselves make a benchmark more realistic or more useful, and the framework offers controllable heterogeneity along several dimensions that need not resemble any particular deployment. Our evidence on what geometric partitions separate covers content and scene but not acquisition factors such as device, sensor, and compression, which would require a representation function sensitive to them. The semantic partitioning pipeline depends on the quality and consistency of VLM and LLM outputs, which may introduce noise, and it is weakest on enumeration criteria, with an NMI of 0.297 for count against 0.52 to 0.86 for the other Clevr-4 criteria, a gap we attribute to the weakness of current VLMs at counting.

\begin{ack}
This work was supported by free cloud compute credits that RunPod provided to Giacomo Tarroni and by the Hyperion HPC cluster of City St George's, University of London. Vasilis Siomos is a full-time employee of Ultromics and Jonathan Passerat-Palmbach is employed by Flashbots, and neither company had any involvement in this work. The other authors declare no competing interests.
\end{ack}

\clearpage
\bibliography{example}
\clearpage
\appendix

\section{The \texttt{fds} package}\label{app:fds}
\texttt{fds}, available at \url{https://github.com/siomvas/fds}, is a Python package for controlled federated learning data simulation, partition evaluation, and FL training. It was designed to make the client-partitioning process explicit, reproducible, and measurable, while also providing the training infrastructure required to study how different forms of data heterogeneity affect FL convergence and final model quality.

\subsection*{Core Capabilities}

\begin{itemize}[leftmargin=*]
    \item \textbf{Federated data simulation:} \texttt{fds} generates client-level datasets for FL experiments under IID, label-skewed, embedding-based, and VLM-informed heterogeneous settings.

    \item \textbf{Reusable partition artifacts:} Client splits are serialized as durable artifacts, allowing the same partition to be inspected, scored, reused for training, and evaluated post hoc.

    \item \textbf{Multi-task support:} The package supports image classification, semantic segmentation, and object detection workflows through task-specific data modules, models, training loops, and evaluation metrics.

    \item \textbf{End-to-end FL execution:} Beyond data splitting, \texttt{fds} can launch full federated training jobs, construct client dataloaders, instantiate models, aggregate client updates, checkpoint global models, and evaluate final models.

    \item \textbf{Experiment analysis:} The package includes tools for quantifying partition heterogeneity, summarizing experiment grids, producing reports, and comparing convergence behavior across partitioning methods.
\end{itemize}

\subsection*{Datasets and Tasks}
Datasets are normalized into sample dictionaries. Classification samples contain image and label fields, segmentation samples contain image and mask references, and detection samples contain image annotations such as boxes, labels, and image identifiers. Federated partitions are represented as mappings from client identifiers to sample indices. This decouples data partitioning from model training and makes each partition independently reproducible. The package supports shared global validation, client-local validation, and no-validation training. This allows experiments to distinguish between during-training validation and post-hoc evaluation of the final global model. Splitters are seed-controlled, enabling repeated generation of the same client partitions when the same configuration is used. Saved split files can be reused across model architectures, FL optimizers, or evaluation scripts without regenerating the underlying client assignments.

The package includes an easily extensible dataset factory, which currently supports classification datasets such as Tiny-ImageNet and Food101, segmentation datasets such as ADE20K, and objection detection datasets such as Pascal VOC and BDD100K. Dataset modules apply appropriate preprocessing for each task, including classification image normalization and augmentation, segmentation image-mask transforms, and detection-specific annotation handling. The same data-loading machinery can support centralized baselines and federated client-specific training. \texttt{fds} also includes a model factory that instantiates classification, segmentation, and detection models from supported providers \texttt{timm} and \texttt{torchvision}, \texttt{transformers} and \texttt{segmentation-models-pytorch} backbones.

\subsection*{Partitioning Methods}

\textbf{Random splitting.} The random splitter provides IID or near-IID baselines by distributing samples approximately evenly across clients.

\textbf{Label-Dirichlet splitting.} The Dirichlet splitter creates label-skewed non-IID clients. The concentration parameter $\alpha$ controls the severity of the label imbalance, with smaller values producing more heterogeneous client label distributions.

\textbf{Embedding-based splitting.} Embedding-based partitioning constructs clients according to similarity in feature space rather than according to labels alone.

\textbf{VLM-based multivariate splitting.} The multivariate splitter uses semantic attributes derived from vision-language annotations, allowing client heterogeneity to vary along interpretable axes such as object type, color, size, scene, location, lighting, or viewpoint.

\textbf{Feature Extractors.} The embedding pipeline supports representations from models such as DINOv2, CLIP, SigLIP, Inception-style networks, MAE-style encoders, raw-pixel projections, and lightweight autoencoder representations. Four registered representation functions use no pretrained weights, namely a raw-pixel projection, a sparse random projection, an untrained autoencoder, and a wavelet scattering transform. For transformer-style backbones, \texttt{fds} supports different token-selection and pooling strategies, such as class-token features, patch-token aggregation, and pooled multi-layer representations. Feature normalization and dimensionality reduction, including UMAP, PCA, and sliced-Wasserstein projections, can be applied before clustering, allowing the user to study how representation geometry affects client construction. Embedding partitions can be generated using methods such as K-means, HDBSCAN, KNN-based grouping, and spectral clustering.

\textbf{Embedding cache.} The package also includes a cache manager. Extracted features are cached, so expensive embedding computation can be reused across partitioning methods, client counts, and hyper-parameter searches.

\subsection*{Partition Evaluation Metrics}

\textbf{Label-distribution metrics:} \texttt{fds} computes class-balance and label-divergence statistics to quantify label skew across clients.

\textbf{Feature-distribution metrics:} The package also evaluates heterogeneity in embedding space using distance-based and distributional measures like FID, MMD, feature-space energy distance, sliced Wasserstein distance, PCA-based multivariate KS diagnostics, kNN separability, and centroid distance.

\textbf{Client balance diagnostics:} The package records sample-count imbalance and related diagnostics so that pathological or undersized client splits can be identified.

\subsection*{Federated Training Infrastructure}

\texttt{fds} uses NVIDIA FLARE~\cite{nvflare} to launch and coordinate federated training jobs. Experiments are configured using structured Hydra configs that specify the dataset, partition file, model, optimizer, scheduler, FL method, client count, number of local epochs, validation policy, and output directory. Each client reconstructs its local dataloader from the saved split file and trains on its assigned samples for the configured number of local epochs. The server saves global model states during training, enabling later post-hoc evaluation or offline continuation. Runs can emit logs, manifests, TensorBoard metrics, validation summaries, and model checkpoints for downstream analysis. The comparison of aggregation methods in Section~\ref{sec:algos} instead runs the implementations of the fluke library~\citep{fluke} through an adapter in \texttt{fds} (Appendix~\ref{app:hparam}).

\subsection*{Released Artifacts}
The release includes the precomputed criterion labels of the four datasets, the prompts and label sets of Appendix~\ref{app:vlm}, and the artifacts and configurations of the Clevr-4 and Food-4c experiments of Section~\ref{sec:validation}.

\section{Hyperparameter details}\label{app:hparam}
All federated learning experiments used FedAvg \cite{fedavg} with full client participation in every communication round. We train for one local epoch per round using mixed-precision training and resumed optimizer state across rounds. Experiments were trained for 100 communication rounds. Validation was performed on the global validation set during training.

For Tiny-ImageNet, we used an ImageNet-pretrained \texttt{timm} \cite{rw2019timm} ResNet-18 classifier with 200 output classes, cross-entropy loss, batch size 32, SGD with learning rate 0.01, momentum 0.9, weight decay 5e-6, and a cosine learning-rate schedule with minimum learning rate 1e-4. For Food101, we used 101 classes, cross-entropy loss, batch size 64, SGD with learning rate 0.01, momentum 0.9, weight decay 0, and no scheduler. For Pascal VOC, we used a COCO-pretrained \texttt{torchvision} Faster R-CNN ResNet50-FPN-v2 detector with 21 classes including background, batch size 2, AdamW with learning rate 1.10e-5, weight decay 0, and no scheduler. For ADE20K, we used a COCO/VOC-pretrained DeepLabV3-ResNet50 segmentation model with 150 classes and Dice loss, batch size 2, AdamW \cite{adamw} with learning rate 4.5e-5, weight decay 1e-4, and a cosine schedule.

Model hyperparameters were selected with Optuna \cite{akiba2019optuna} using a single-client proxy training setup and validation objectives matched to each task: accuracy for classification, mean IoU for segmentation, and mAP@0.5 for detection. We use TPE sampling with median pruning and search over batch size, optimizer, scheduler, log-scaled learning rate, weight decay, SGD momentum, and the pretrained flag for classification backbones. Separately, embedding-based partition selection was tuned with Optuna performing a full grid search over the embedding extractor, token/pooling strategy, normalization, dimensionality reduction using MMD-based heterogeneity scoring.

\textbf{Comparison of aggregation methods.} The experiment of Section~\ref{sec:algos} uses Tiny ImageNet at its native resolution of $64\times64$ with 10 clients, a ResNet-18 with group normalization trained from random initialization and an ImageNet-pretrained ResNet-18 with batch normalization, and 100 rounds of one local epoch with all clients participating, with a batch size of 64, SGD with a learning rate of 0.01, momentum 0.9, no weight decay, and no scheduler, and a server learning rate of 1.0. The methods are the implementations of fluke~\citep{fluke} (version 0.8.1), run through \texttt{fds}, which adapts the fluke implementations of FedDyn, FedNova, and SCAFFOLD. Each configuration is run with three seeds on the same partitions. The batch size, input resolution, weight decay, and schedule differ from those of the main Tiny ImageNet experiments, which accounts for the lower accuracies of Table~\ref{tab:algos}.

\section{Semantic Clustering Pipeline Details}\label{app:vlm}
The semantic partitioning pipeline follows \texttt{IC|TC} \cite{ictc}, with the key differences being modified prompts and additional guardrails.
\subsection{Step 1: Criterion-Focused Image Description}

Step 1 takes each image and asks a VLM to describe the image while focusing on a single criterion. The selected criterion is substituted into the dataset-specific \texttt{step1\_prompt.txt}. The output for each image is a record containing the image path, original dataset label, generated description, and original sample index. The exact Step 1 prompt template is:

\begin{verbatim}
Describe the image focusing exclusively on [__CRITERION__]. 
Avoid proper nouns. Answer in <=80 tokens.
\end{verbatim}

For example, Tiny-ImageNet \texttt{main\_object} uses the prompt label  \texttt{Main Object}, while Food101 \texttt{cuisine} uses \texttt{cuisine or culinary origin}.

\subsection{Step 2a: Raw Criterion Label Extraction}

Step 2a takes the Step 1 description and asks an LLM to produce a one-word answer for the selected criterion. For example, for Tiny-ImageNet, the exact Step 2a system prompt template is:

\begin{verbatim}
You will be given a description of an image. Your job is 
to determine the [__CRITERION__] 
within the image based on the provided description. 
Please respond in a single word. 
For example, given the following description:

"""
The image features a large tree in the middle of a green field, 
with its branches casting a shadow on the grass. The tree 
appears a willow tree, and its branches are covered in green leaves. 
The sun is shining, creating a beautiful morning scene.
"""
An exemplar answer is "[__EXEMPLAR__]".
\end{verbatim}
The exact Step 2a user prompt generated for each caption is:

\begin{verbatim}
Task: Read the image description and respond 
with one word about the "{args.prompt_label}".
Image description: "{caption}" 
Your response:
\end{verbatim}
Step 2a writes the LLM output into the \texttt{label} field of each record in \texttt{step2a\_result.jsonl}. These labels are raw, unconstrained strings and may include synonyms, spelling variants, or overly specific words.

\subsection{Step 2b: Ontology Induction from Raw Labels}

Step 2b takes all the raw labels, removes punctuation, counts label frequencies, filters low-frequency labels, and asks the LLM to cluster the remaining labels into a requested number of ontology categories. The exact Step 2b system prompt template, shared across datasets, is:

\begin{verbatim}
You will be provided a list of [__LEN__] [__CRITERION__] 
and the number of occurences in a given dataset. 
Your job is to cluster [__LEN__] words into 
[__NUM_CLASSES_CLUSTER__] categories. 
Provide your answer as a list of [__NUM_CLASSES_CLUSTER__] words, 
each word representing a [__CRITERION__].

You must provide your answer in the following format 
"Answer {index}: {object}", where {index} is the 
index of the category and {object} is the object name 
representing the category. For example, if you think the first category 
is "object", then you should provide your answer as "Answer 1: object".

When categorizing classes, consider the following criteria:

1. Merge clusters with similar meanings.
2. Each cluster represents a general [__CRITERION__] 
and should not be too specific.

Also, please provide a reason you chose the word for each category. 
You can provide your reason in the following format 
"Reason {index}: {reason}", where {index} is the index of the 
category and {reason} is the reason you chose the word for the category.
\end{verbatim}

The exact Step 2b user prompt generated by \texttt{step2b.format\_prompt} is:

\begin{verbatim}
Task: Cluster these following list of labels into 
{num_of_class} categories.
List of labels: {labels}
Your response:
\end{verbatim}

\subsection{Step 3: Final Constrained Classification}

The final stage reloads the Step 1 image descriptions and the Step 2b ontology labels, then asks the LLM to assign each image description to exactly one ontology label. This is the annotation file consumed by the VLM partition splitter. The exact Step 3 system prompt template is:

\begin{verbatim}
You are a classification system. 
Your task: Based on the image description, determine the most 
appropriate category that best classifies the [__CRITERION__] in the image. 
You must choose a category from this list: [__CLASSES__]
IMPORTANT RULES:
- Output only the category name
- No explanations or extra text
- Choose the most specific matching category
- If ambiguous, select the closest match based on dominant feature, 
no explanations needed
- If multiple categories could apply, choose the most dominant one
- Format: CategoryName (exactly as provided in options)
\end{verbatim}

The exact Step 3 user prompt generated by \texttt{step3.format\_prompt} is:

\begin{verbatim}
Task: Based on the image description, determine the category for 
the {args.prompt_label} in the image. 
You must choose one option from this list: {class_list}

Image description: "{caption}".

Base your decision on the primary/dominant characteristic. 
If a situation arises where nothing is allocated, 
please assign it to the object that has the closest resemblance. 

Output only the category name.
Your response:
\end{verbatim}

\subsection{Selected Criteria}

\begin{description}
    \item[\textbf{Tiny-ImageNet:}] \texttt{main\_object}, \texttt{size}, \texttt{time}, \texttt{color}, \texttt{location}.
    \item[\textbf{Food101:}] \texttt{cuisine}, \texttt{preparation}, \texttt{composition}, \texttt{texture}, \texttt{formality}.
    \item[\textbf{ADE20K:}] \texttt{main\_object}, \texttt{scene}, \texttt{color}, \texttt{light}, \texttt{size}, \texttt{camera}.
    \item[\textbf{PASCAL:}] \texttt{main\_object}, \texttt{location}, \texttt{color}, \texttt{light}, \texttt{size}.
\end{description}

\subsection{Criterion Prompt Labels, Exemplars, and Requested Ontology Sizes}

\begin{center}
\small
\begin{tabular}{llll}
\hline
Dataset & Criterion & Prompt label & Exemplar / requested $K$ \\
\hline
\texttt{tiny} & \texttt{main\_object} & Main Object & Tree / 7 \\
\texttt{tiny} & \texttt{size} & Size and Scale of the main object & Large / 4 \\
\texttt{tiny} & \texttt{time} & Time of day & Morning / 4 \\
\texttt{tiny} & \texttt{color} & Dominant Color & Green / 7 \\
\texttt{tiny} & \texttt{location} & Location & Field / 5 \\
\hline
\texttt{food101} & \texttt{cuisine} & cuisine or culinary origin & Italian / 5 \\
\texttt{food101} & \texttt{preparation} & cooking or preparation method & fried / 5 \\
\texttt{food101} & \texttt{composition} & arrangement and structure of food components & mixed / 4 \\
\texttt{food101} & \texttt{texture} & dominant surface texture & crispy / 4 \\
\texttt{food101} & \texttt{formality} & plating style and presentation formality & formal / 3 \\
\hline
\texttt{ade20k} & \texttt{main\_object} & Main Object & Tree / 8 \\
\texttt{ade20k} & \texttt{scene} & Location & Field / 4 \\
\texttt{ade20k} & \texttt{color} & Dominant Color & Green / 7 \\
\texttt{ade20k} & \texttt{light} & Lighting Intensity & Bright / 4 \\
\texttt{ade20k} & \texttt{size} & Size of the main object & Large / 5 \\
\texttt{ade20k} & \texttt{camera} & Camera Perspective & Wide-shot / 5 \\
\hline
\texttt{PascalVOC} & \texttt{main\_object} & Main Object & Tree / 5 \\
\texttt{PascalVOC} & \texttt{location} & Location & Field / 4 \\
\texttt{PascalVOC} & \texttt{color} & Dominant Color & Green / 6 \\
\texttt{PascalVOC} & \texttt{light} & Lighting Intensity & Bright / 4 \\
\texttt{PascalVOC} & \texttt{size} & Size of the main object & Large / 5 \\
\hline
\end{tabular}
\end{center}

\subsection{Final Label Ontology per Criterion}

The following label lists are the final ontology labels extracted from the saved training-split \texttt{step2b\_result.txt} files. These labels are the candidate classes supplied to the final constrained classification stage.

\subsubsection*{Tiny-ImageNet}

\begin{description}
    \item[\texttt{color}] \texttt{Brown/Yellow}, \texttt{Red}, \texttt{Green}, \texttt{Blue}, \texttt{Black}, \texttt{White}, \texttt{Otherd}.
    \item[\texttt{location}] \texttt{Indoors}, \texttt{Urban Outdoors}, \texttt{Nature Outdoors}, \texttt{Aquatic}, \texttt{Unknown}.
    \item[\texttt{main\_object}] \texttt{Human}, \texttt{Animal}, \texttt{Vehicle}, \texttt{Building}, \texttt{Food}, \texttt{Landscape}, \texttt{Inanimate Object}.
    \item[\texttt{size}] \texttt{Small}, \texttt{Large}, \texttt{Medium}, \texttt{Expansive}.
    \item[\texttt{time}] \texttt{Day}, \texttt{Night}, \texttt{Unknown}, \texttt{Sunset/Dusk}.
\end{description}

\subsubsection*{Food101}

\begin{description}
    \item[\texttt{composition}] \texttt{Mixed}, \texttt{Layered/Stacked}, \texttt{Scattered/Spread}, \texttt{Uniform}.
    \item[\texttt{cuisine}] \texttt{American}, \texttt{Italian}, \texttt{Asian}, \texttt{Mexican}, \texttt{French}.
    \item[\texttt{formality}] \texttt{Casual}, \texttt{Elegant}, \texttt{Modern}.
    \item[\texttt{preparation}] \texttt{Baked}, \texttt{Fried}, \texttt{Steamed}, \texttt{Raw}, \texttt{Mixed}.
    \item[\texttt{texture}] \texttt{Smooth}, \texttt{Rough}, \texttt{Crispy}, \texttt{Soft}.
\end{description}

\subsubsection*{ADE20K}

\begin{description}
    \item[\texttt{camera}] \texttt{Lowangle}, \texttt{Eyelevel}, \texttt{Highangle}, \texttt{Aerial}, \texttt{Wideangle}.
    \item[\texttt{color}] \texttt{Brown/Beige}, \texttt{White}, \texttt{Green}, \texttt{Gray}, \texttt{Blue}, \texttt{Red/Orange}, \texttt{Black}.
    \item[\texttt{light}] \texttt{Intense}, \texttt{Moderate}, \texttt{Dim}.
    \item[\texttt{main\_object}] \texttt{Building}, \texttt{Furniture}, \texttt{Vehicle}, \texttt{Person}, \texttt{Nature}.
    \item[\texttt{scene}] \texttt{Urban Outdoors}, \texttt{Residential Indoors}, \texttt{Natural Outdoors}, \texttt{Office Indoors}.
    \item[\texttt{size}] \texttt{Large}, \texttt{Small}, \texttt{Medium}, \texttt{Huge}.
\end{description}

\subsubsection*{Pascal VOC}

\begin{description}
    \item[\texttt{color}] \texttt{Red/Orange}, \texttt{Blue/Purple}, \texttt{Green}, \texttt{Black}, \texttt{Brown/Yellow}, \texttt{White}.
    \item[\texttt{light}] \texttt{Dim}, \texttt{Moderate}, \texttt{Bright}, \texttt{Intense}.
    \item[\texttt{location}] \texttt{Indoors}, \texttt{Urban Outdoors}, \texttt{Nature}.
    \item[\texttt{main\_object}] \texttt{Animal}, \texttt{Person}, \texttt{Vehicle}, \texttt{Nature}, \texttt{Unanimated Object}, \texttt{Building}.
    \item[\texttt{size}] \texttt{Large}, \texttt{Small}, \texttt{Medium}, \texttt{Huge}.
\end{description}

\section{Pure-Criterion Ablation}\label{app:ablation}
Table~\ref{tab:criterion-ablation} gives the feature-space heterogeneity of the pure-criterion splits of Section~\ref{sec:criterion-ablation}.

\begin{table}[t]
\centering
\caption{
Pure-criterion ablation for semantic partitioning. For each criterion, one client is created per criterion value and all samples with that value are assigned to the corresponding client. Higher values indicate stronger feature-space separation.
}
\label{tab:criterion-ablation}
\resizebox{0.75\linewidth}{!}{
\begin{tabular}{llrrrrr}
\toprule
Dataset & Criterion & Clients & FID $\uparrow$ & MMD $\uparrow$ & $k$NN sep. $\uparrow$ & Centroid $\uparrow$ \\
\midrule
Tiny ImageNet & main object & 7 & \textbf{115.3} & \textbf{0.0368} & \textbf{0.840} & 13.7 \\
 & location & 5 & 73.2 & 0.0237 & 0.781 & 10.9 \\
 & size & 4 & 60.2 & 0.0160 & 0.621 & 8.6 \\
 & color & 7 & 41.3 & 0.0092 & 0.482 & 7.0 \\
 & time & 4 & 29.8 & 0.0073 & 0.583 & 6.2 \\
\midrule
ADE20K & scene & 4 & 122.1 & \textbf{0.0396} & \textbf{0.853} & 14.2 \\
 & main object & 5 & 118.4 & 0.0275 & 0.761 & 11.6 \\
 & camera & 5 & 111.2 & 0.0259 & 0.567 & 11.0 \\
 & color & 7 & \textbf{165.0} & 0.0197 & 0.632 & 10.4 \\
 & size & 4 & 47.7 & 0.0093 & 0.538 & 6.9 \\
 & light & 3 & 17.8 & 0.0028 & 0.465 & 3.8 \\
\midrule
PASCAL VOC & main object & 6 & \textbf{228.5} & \textbf{0.0358} & \textbf{0.851} & 14.1 \\
 & location & 3 & 97.8 & 0.0182 & 0.729 & 9.8 \\
 & size & 4 & 90.0 & 0.0131 & 0.515 & 8.0 \\
 & color & 6 & 89.1 & 0.0069 & 0.329 & 6.3 \\
 & light & 4 & 57.9 & 0.0035 & 0.339 & 4.5 \\
\midrule
Food101 & cuisine & 5 & 34.7 & \textbf{0.0616} & \textbf{0.764} & 16.5 \\
 & preparation & 5 & \textbf{42.4} & 0.0584 & 0.740 & 16.1 \\
 & composition & 4 & 34.5 & 0.0274 & 0.582 & 11.3 \\
 & texture & 4 & 22.5 & 0.0231 & 0.629 & 10.0 \\
 & formality & 3 & 16.9 & 0.0212 & 0.733 & 9.8 \\
\bottomrule
\end{tabular}}
\end{table}

\section{Sensitivity of Geometric Partitioning to the Representation}\label{app:sensitivity}
Geometric partitioning involves several choices beyond the encoder, and we measured how much the representation function, normalization, and dimensionality reduction each matter. Over 733 configurations at $K = 10$, spanning the four datasets, eight representation functions, and reduction by UMAP to 32 or 128 dimensions or no reduction, the representation function accounts for between 58.8\% and 76.3\% of the variance in MMD depending on the dataset. The reduction accounts for between 0.02\% and 7.8\% and normalization for less than 0.2\%, and FID and centroid distance order the factors in the same way. UMAP raises MMD by a median factor of 1.39 on Food101 and 1.08 on Tiny ImageNet, has no effect on the two dense prediction datasets, and ranks configurations as the unreduced embeddings do, with Spearman correlations between 0.885 and 0.909.

A further sweep of 160 configurations crosses the four datasets with four encoders and ten reduction settings, including PCA to 32 and 128 dimensions and variants of the UMAP parameters. PCA never changes which encoder produces the most heterogeneous partition, although at 128 components it shifts the magnitude of MMD by $+17\%$ on ADE20K and $-22\%$ on Tiny ImageNet. DINOv2-B, selected for the main experiments by its average MMD rank, remains the encoder with the highest MMD under the UMAP settings deployed there. Other UMAP settings change the leading encoder in six arms on Tiny ImageNet and Food101, while DINOv2-B leads throughout on ADE20K and PASCAL VOC.

\section{Semantic Pipeline: Validation, Models, and Compute}\label{app:validation}
\textbf{Clevr-4.} Clevr-4~\citep{vaze2023norep} consists of synthetic scenes that admit four equally valid taxonomies, by shape, texture, color, and count. We run the pipeline with InternVL3.5-38B and Llama 3.3-70B on 10,531 Clevr-4 images and compare the partition produced for each criterion with each of the four taxonomies (Table~\ref{tab:clevr-matrix}). No stage of the pipeline reads an annotation, but the texture prompt was revised once during the run, after the descriptions produced so far had been clustered and scored against the texture taxonomy. Measured by AMI, each criterion agrees with its own taxonomy far more than with any other, by factors of 277 for shape, 34 for color, 21 for count, and 8.5 for texture over the strongest other taxonomy, and the AMI off the diagonal averages 0.010. Two controls bound these values. The four annotated taxonomies are nearly independent of one another, with a mean AMI of 0.0007, and the four criterion partitions share little with one another, with a mean AMI of 0.012. Count, the one enumeration criterion, is recovered least completely, which we attribute to the weakness of current vision-language models at counting. With Qwen3-VL-8B~\citep{qwen3vl} and Qwen3-8B~\citep{qwen3} in place of the two models, every criterion still peaks on its own taxonomy, while the NMI falls from 0.860 to 0.674 for shape and from 0.793 to 0.711 for color, so the capability of the models governs how completely a criterion is recovered and not which axis is recovered. This run kept the original texture prompt, so its texture NMI of 0.272 does not isolate the effect of the models.

\begin{table}[h]
\centering
\caption{Clevr-4 validation of the semantic pipeline (InternVL3.5-38B and Llama 3.3-70B). (a) NMI between the partition produced for each criterion and each annotated taxonomy. (b) AMI between the four criterion partitions. (c) AMI between the four annotated taxonomies.}
\label{tab:clevr-matrix}
\small
\begin{tabular}{@{}lcccc@{}}
\multicolumn{5}{@{}l}{(a) Criterion (rows) against taxonomy (columns), NMI} \\
\toprule
 & Texture & Shape & Color & Count \\
\midrule
Texture & \textbf{0.524} & 0.063 & 0.012 & 0.003 \\
Shape & 0.004 & \textbf{0.860} & 0.002 & 0.005 \\
Color & 0.025 & 0.008 & \textbf{0.793} & 0.002 \\
Count & 0.002 & 0.016 & 0.003 & \textbf{0.297} \\
\bottomrule
\end{tabular}
\\[6pt]
\begin{tabular}{@{}lcccc@{}}
\multicolumn{5}{@{}l}{(b) Criterion against criterion, AMI} \\
\toprule
 & Texture & Shape & Color & Count \\
\midrule
Texture & -- & 0.041 & 0.019 & 0.001 \\
Shape & 0.041 & -- & 0.005 & 0.008 \\
Color & 0.019 & 0.005 & -- & 0.001 \\
Count & 0.001 & 0.008 & 0.001 & -- \\
\bottomrule
\end{tabular}\quad
\begin{tabular}{@{}lcccc@{}}
\multicolumn{5}{@{}l}{(c) Taxonomy against taxonomy, AMI} \\
\toprule
 & Texture & Shape & Color & Count \\
\midrule
Texture & -- & 0.0000 & -0.0001 & 0.0001 \\
Shape & 0.0000 & -- & 0.0000 & 0.0043 \\
Color & -0.0001 & 0.0000 & -- & -0.0001 \\
Count & 0.0001 & 0.0043 & -0.0001 & -- \\
\bottomrule
\end{tabular}
\end{table}

\textbf{Food-4c.} Food-4c~\citep{liu2026organizing} supplies human annotations of Food101 images along four criteria, cuisine among them. On 68,250 Food101 images, with the human cuisine labels coarsened to five categories, the cuisine labels of our pipeline reach an adjusted mutual information (AMI) of 0.503 with the human labels, an adjusted Rand index of 0.465, and direct agreement on 73.0\% of images. Pipeline criteria that do not correspond to cuisine score far lower against the same human labels, 0.092 for preparation and 0.018 for texture, as do the pipeline's cuisine labels against an unrelated human criterion, at 0.095. The five criteria of the pipeline are also nearly independent of one another, with a mean pairwise AMI of 0.032 and a maximum of 0.071, whereas the human taxonomies of Food-4c overlap far more, with a mean pairwise AMI of 0.360 and a maximum of 0.625.

\textbf{Models.} We tried the vision-language models MiniCPM-V 2.6~\citep{minicpmv}, InternVL2~\citep{internvl}, InternVL2.5-38B~\citep{internvl25}, InternVL3.5 at 8B and 38B parameters~\citep{internvl3}, and Qwen3-VL at 8B and 30B parameters~\citep{qwen3vl}, and the language models Llama 3.1-8B-Instruct and Llama 3.3-70B~\citep{llama}. Larger models and newer generations performed better, and above a threshold of capability the choice of model family mattered little.

\textbf{Compute.} The labeling pipeline ran on two A100 GPUs with 80\,GB of memory each. Labeling took about three days per criterion on the classification datasets of roughly 100,000 images and less than a day per criterion on ADE20K and PASCAL VOC, with criteria labeled concurrently. Labeling the 10,531 Clevr-4 images for all four criteria took about 16 hours and 4.5 million tokens.

\section{Client Composition of Geometric Partitions}\label{app:composition}
We compare each geometric partition of the heterogeneity tables with the pure-criterion partition of each criterion, for the four datasets and $K \in \{5, 10, 20\}$, which gives 63 combinations of dataset, criterion, and client count (Table~\ref{tab:composition-ami}). The criterion labels cover every sample of each partition (100,000 on Tiny ImageNet, 75,750 on Food101, 25,574 on ADE20K, and 5,717 on PASCAL VOC). Over all datasets, the content criteria (main object, scene, and cuisine) reach a mean AMI of 0.379, 0.377, and 0.363 at $K = 5$, 10, and 20, against 0.056, 0.060, and 0.061 for lighting, color, camera, and size. On ADE20K at $K = 10$ the ordering runs from scene (0.417) and main object (0.334) through camera (0.095), size (0.078), and color (0.033) to lighting (0.013), and total variation distances agree. Alignment also shifts with the number of clients, since on ADE20K alignment with scene falls from 0.492 at $K = 5$ to 0.382 at $K = 20$, whereas on Food101 alignment with cuisine rises from 0.117 to 0.255.

Individual clients carry recognizable signatures (Table~\ref{tab:signatures}). On Tiny ImageNet with 10 clients, one client holds vehicles at 10.6 times their global rate together with urban outdoor scenes at 4.3 times, another food at 9.8 times with indoor scenes at 2.1 times, and a third humans at 4.7 times. On ADE20K with five clients, one client consists of 96.1\% residential indoor scenes against 31.7\% in the dataset as a whole.

\begin{table}[h]
\centering
\caption{AMI between each geometric partition and the pure-criterion partition of each criterion, for $K \in \{5, 10, 20\}$. $^{c}$Content criteria. $^{s}$Criteria of lighting, color, camera, and size.}
\label{tab:composition-ami}
\small
\begin{tabular}{@{}llccc@{}}
\toprule
Dataset & Criterion & $K=5$ & $K=10$ & $K=20$ \\
\midrule
Tiny ImageNet & main object$^{c}$ & 0.483 & 0.515 & 0.486 \\
 & size$^{s}$ & 0.073 & 0.084 & 0.085 \\
 & time & 0.080 & 0.081 & 0.075 \\
 & color$^{s}$ & 0.053 & 0.059 & 0.072 \\
 & location & 0.241 & 0.237 & 0.233 \\
\midrule
Food101 & cuisine$^{c}$ & 0.117 & 0.148 & 0.255 \\
 & preparation & 0.146 & 0.219 & 0.257 \\
 & composition & 0.064 & 0.067 & 0.090 \\
 & texture & 0.025 & 0.050 & 0.061 \\
 & formality & 0.038 & 0.050 & 0.055 \\
\midrule
ADE20K & main object$^{c}$ & 0.372 & 0.334 & 0.309 \\
 & scene$^{c}$ & 0.492 & 0.417 & 0.382 \\
 & color$^{s}$ & 0.036 & 0.033 & 0.031 \\
 & light$^{s}$ & 0.007 & 0.013 & 0.014 \\
 & size$^{s}$ & 0.083 & 0.078 & 0.078 \\
 & camera$^{s}$ & 0.107 & 0.095 & 0.087 \\
\midrule
PASCAL VOC & main object$^{c}$ & 0.432 & 0.472 & 0.381 \\
 & location & 0.111 & 0.226 & 0.190 \\
 & color$^{s}$ & 0.041 & 0.052 & 0.060 \\
 & light$^{s}$ & 0.013 & 0.028 & 0.032 \\
 & size$^{s}$ & 0.086 & 0.096 & 0.094 \\
\bottomrule
\end{tabular}
\end{table}

\begin{table}[h]
\centering
\caption{Client signatures of the geometric partitions at $K = 10$: for each client, the three criterion values with the largest lift (client rate over global rate) among values covering at least 5\% of the client.}
\label{tab:signatures}
\scriptsize
\begin{tabular}{@{}llp{0.78\linewidth}@{}}
\toprule
Dataset & Client & Signature \\
\midrule
Tiny ImageNet & 0 & location=Aquatic (3.66$\times$), main object=Animal (2.65$\times$), size=Small (1.79$\times$) \\
 & 1 & main object=Building (6.44$\times$), size=Expansive (5.70$\times$), main object=Landscape (5.33$\times$) \\
 & 2 & main object=Inanimate Object (2.73$\times$), time=Unknown (1.98$\times$), time=Night (1.87$\times$) \\
 & 3 & main object=Animal (3.53$\times$), location=Nature Outdoors (2.60$\times$), color=Brown/Yellow (1.97$\times$) \\
 & 4 & main object=Vehicle (10.55$\times$), location=Urban Outdoors (4.34$\times$), color=White (2.01$\times$) \\
 & 5 & main object=Human (4.68$\times$), size=Medium (1.86$\times$), time=Unknown (1.74$\times$) \\
 & 6 & main object=Food (9.84$\times$), location=Indoors (2.06$\times$), color=Red (1.97$\times$) \\
 & 7 & main object=Animal (3.62$\times$), location=Aquatic (3.24$\times$), location=Nature Outdoors (2.33$\times$) \\
 & 8 & color=Black (4.75$\times$), main object=Animal (3.51$\times$), location=Nature Outdoors (2.44$\times$) \\
 & 9 & main object=Animal (3.51$\times$), size=Medium (1.49$\times$), location=Indoors (1.44$\times$) \\
\midrule
Food101 & 0 & preparation=Raw (1.86$\times$), formality=Modern (1.75$\times$), formality=Elegant (1.56$\times$) \\
 & 1 & cuisine=Mexican (2.50$\times$), preparation=Mixed (1.86$\times$), preparation=Baked (1.38$\times$) \\
 & 2 & texture=Crispy (3.31$\times$), preparation=Fried (3.01$\times$), composition=Scattered/Spread (1.68$\times$) \\
 & 3 & preparation=Baked (2.29$\times$), composition=Uniform (1.84$\times$), cuisine=French (1.66$\times$) \\
 & 4 & composition=Uniform (6.82$\times$), cuisine=American (3.36$\times$), preparation=Raw (2.67$\times$) \\
 & 5 & preparation=Mixed (5.77$\times$), cuisine=Italian (5.52$\times$), composition=Mixed (3.84$\times$) \\
 & 6 & preparation=Steamed (4.92$\times$), cuisine=French (3.37$\times$), preparation=Raw (2.28$\times$) \\
 & 7 & preparation=Steamed (9.82$\times$), cuisine=Asian (4.55$\times$), composition=Mixed (2.47$\times$) \\
 & 8 & cuisine=Mexican (2.12$\times$), texture=Crispy (1.76$\times$), texture=Soft (1.67$\times$) \\
 & 9 & preparation=Steamed (10.50$\times$), cuisine=Asian (4.64$\times$), composition=Scattered/Spread (2.41$\times$) \\
\midrule
ADE20K & 0 & main object=Person (2.27$\times$), scene=Natural Outdoors (2.02$\times$), camera=Highangle (1.71$\times$) \\
 & 1 & scene=Residential Indoors (3.14$\times$), main object=Furniture (3.02$\times$), color=Red/Orange (2.65$\times$) \\
 & 2 & camera=Wideangle (3.44$\times$), color=Black (2.79$\times$), main object=Furniture (2.73$\times$) \\
 & 3 & scene=Residential Indoors (3.03$\times$), main object=Furniture (2.75$\times$), size=Medium (2.62$\times$) \\
 & 4 & scene=Urban Outdoors (2.77$\times$), main object=Vehicle (2.26$\times$), main object=Building (2.26$\times$) \\
 & 5 & main object=Nature (3.82$\times$), scene=Natural Outdoors (2.98$\times$), size=Small (1.85$\times$) \\
 & 6 & scene=Office Indoors (4.15$\times$), color=Black (2.74$\times$), main object=Person (2.49$\times$) \\
 & 7 & scene=Residential Indoors (2.78$\times$), camera=Wideangle (2.67$\times$), main object=Furniture (2.42$\times$) \\
 & 8 & scene=Urban Outdoors (2.58$\times$), main object=Building (2.26$\times$), main object=Vehicle (1.60$\times$) \\
 & 9 & camera=Wideangle (3.13$\times$), main object=Furniture (2.85$\times$), scene=Residential Indoors (2.58$\times$) \\
\midrule
PASCAL VOC & 0 & main object=Animal (2.75$\times$), size=Small (2.27$\times$), color=Brown/Yellow (1.41$\times$) \\
 & 1 & main object=Vehicle (2.73$\times$), location=Urban Outdoors (2.15$\times$), color=Red/Orange (1.71$\times$) \\
 & 2 & main object=Vehicle (3.44$\times$), size=Huge (2.65$\times$), color=White (2.23$\times$) \\
 & 3 & color=Green (2.88$\times$), location=Nature (2.69$\times$), main object=Animal (2.64$\times$) \\
 & 4 & main object=Unanimated Object (3.29$\times$), main object=Person (2.36$\times$), location=Indoors (1.99$\times$) \\
 & 5 & size=Huge (5.37$\times$), main object=Vehicle (3.29$\times$), location=Urban Outdoors (2.26$\times$) \\
 & 6 & main object=Animal (2.88$\times$), location=Nature (2.59$\times$), color=Green (1.50$\times$) \\
 & 7 & main object=Animal (2.89$\times$), location=Indoors (2.12$\times$), size=Medium (1.98$\times$) \\
 & 8 & main object=Animal (2.54$\times$), location=Nature (2.44$\times$), size=Large (1.98$\times$) \\
 & 9 & main object=Building (6.07$\times$), size=Huge (2.65$\times$), main object=Vehicle (2.26$\times$) \\
\bottomrule
\end{tabular}
\end{table}

\section{Practical Guidance}\label{app:guidance}
\textbf{Choosing the representation.} Content-level heterogeneity, which Section~\ref{sec:geometry} shows geometric partitions to follow, needs no intervention beyond the default encoder. Other axes call for attention to the biases of the encoder, since objectives that reconstruct masked pixels and objectives that align embeddings capture different properties of an image~\citep{park2023what,vaze2023norep}, and to the projection, removing the top $p$ principal components with $p$ as the control. Where neither suffices, the representation functions of \texttt{fds} that use no pretrained weights (Appendix~\ref{app:fds}), or an encoder trained on the domain in question, can take the place of a general-purpose encoder.

\textbf{Choosing criteria.} To choose criteria, we start by listing the ways in which the images of a dataset actually differ. On Food101, candidate criteria for viewpoint, color, and lighting returned nearly degenerate labels, and none of the three appears among the final criteria (Appendix~\ref{app:vlm}). Each remaining candidate should be validated on a subset for a non-degenerate label distribution, the survivors ranked with the pure-criterion screen of Section~\ref{sec:criterion-ablation}, and the final set checked for redundancy, which on Food101 is low, the five criteria sharing a mean pairwise AMI of 0.032 (Appendix~\ref{app:validation}).


\newpage
\section*{NeurIPS Paper Checklist}

\begin{enumerate}

\item {\bf Claims}
    \item[] Question: Do the main claims made in the abstract and introduction accurately reflect the paper's contributions and scope?
    \item[] Answer: \answerYes{} 
    \item[] Justification: We substantiate the presented claims through our framework in Section 2 and an extensive empirical analysis in Section 3. The code and artifacts are released at \url{https://github.com/siomvas/fds}.
    \item[] Guidelines:
    \begin{itemize}
        \item The answer \answerNA{} means that the abstract and introduction do not include the claims made in the paper.
        \item The abstract and/or introduction should clearly state the claims made, including the contributions made in the paper and important assumptions and limitations. A \answerNo{} or \answerNA{} answer to this question will not be perceived well by the reviewers. 
        \item The claims made should match theoretical and experimental results, and reflect how much the results can be expected to generalize to other settings. 
        \item It is fine to include aspirational goals as motivation as long as it is clear that these goals are not attained by the paper. 
    \end{itemize}

\item {\bf Limitations}
    \item[] Question: Does the paper discuss the limitations of the work performed by the authors?
    \item[] Answer: \answerYes{} 
    \item[] Justification: We discuss the limitation of our work as part of the Conclusion section.
    \item[] Guidelines:
    \begin{itemize}
        \item The answer \answerNA{} means that the paper has no limitation while the answer \answerNo{} means that the paper has limitations, but those are not discussed in the paper. 
        \item The authors are encouraged to create a separate ``Limitations'' section in their paper.
        \item The paper should point out any strong assumptions and how robust the results are to violations of these assumptions (e.g., independence assumptions, noiseless settings, model well-specification, asymptotic approximations only holding locally). The authors should reflect on how these assumptions might be violated in practice and what the implications would be.
        \item The authors should reflect on the scope of the claims made, e.g., if the approach was only tested on a few datasets or with a few runs. In general, empirical results often depend on implicit assumptions, which should be articulated.
        \item The authors should reflect on the factors that influence the performance of the approach. For example, a facial recognition algorithm may perform poorly when image resolution is low or images are taken in low lighting. Or a speech-to-text system might not be used reliably to provide closed captions for online lectures because it fails to handle technical jargon.
        \item The authors should discuss the computational efficiency of the proposed algorithms and how they scale with dataset size.
        \item If applicable, the authors should discuss possible limitations of their approach to address problems of privacy and fairness.
        \item While the authors might fear that complete honesty about limitations might be used by reviewers as grounds for rejection, a worse outcome might be that reviewers discover limitations that aren't acknowledged in the paper. The authors should use their best judgment and recognize that individual actions in favor of transparency play an important role in developing norms that preserve the integrity of the community. Reviewers will be specifically instructed to not penalize honesty concerning limitations.
    \end{itemize}

\item {\bf Theory assumptions and proofs}
    \item[] Question: For each theoretical result, does the paper provide the full set of assumptions and a complete (and correct) proof?
    \item[] Answer: \answerNA{} 
    \item[] Justification: This paper introduces only some definitions relating to the task of federated dataset simulation, which serve to form a common language when making this task the object of study and design. Their relevance and utility is shown empirically in the corresponding section.
    \item[] Guidelines:
    \begin{itemize}
        \item The answer \answerNA{} means that the paper does not include theoretical results. 
        \item All the theorems, formulas, and proofs in the paper should be numbered and cross-referenced.
        \item All assumptions should be clearly stated or referenced in the statement of any theorems.
        \item The proofs can either appear in the main paper or the supplemental material, but if they appear in the supplemental material, the authors are encouraged to provide a short proof sketch to provide intuition. 
        \item Inversely, any informal proof provided in the core of the paper should be complemented by formal proofs provided in appendix or supplemental material.
        \item Theorems and Lemmas that the proof relies upon should be properly referenced. 
    \end{itemize}

    \item {\bf Experimental result reproducibility}
    \item[] Question: Does the paper fully disclose all the information needed to reproduce the main experimental results of the paper to the extent that it affects the main claims and/or conclusions of the paper (regardless of whether the code and data are provided or not)?
    \item[] Answer: \answerYes{} 
    \item[] Justification: We provide a best-effort explanation of all the hyper-parameters used. Along with the released code, this should be enough to replicate our results.
    \item[] Guidelines:
    \begin{itemize}
        \item The answer \answerNA{} means that the paper does not include experiments.
        \item If the paper includes experiments, a \answerNo{} answer to this question will not be perceived well by the reviewers: Making the paper reproducible is important, regardless of whether the code and data are provided or not.
        \item If the contribution is a dataset and\slash or model, the authors should describe the steps taken to make their results reproducible or verifiable. 
        \item Depending on the contribution, reproducibility can be accomplished in various ways. For example, if the contribution is a novel architecture, describing the architecture fully might suffice, or if the contribution is a specific model and empirical evaluation, it may be necessary to either make it possible for others to replicate the model with the same dataset, or provide access to the model. In general. releasing code and data is often one good way to accomplish this, but reproducibility can also be provided via detailed instructions for how to replicate the results, access to a hosted model (e.g., in the case of a large language model), releasing of a model checkpoint, or other means that are appropriate to the research performed.
        \item While NeurIPS does not require releasing code, the conference does require all submissions to provide some reasonable avenue for reproducibility, which may depend on the nature of the contribution. For example
        \begin{enumerate}
            \item If the contribution is primarily a new algorithm, the paper should make it clear how to reproduce that algorithm.
            \item If the contribution is primarily a new model architecture, the paper should describe the architecture clearly and fully.
            \item If the contribution is a new model (e.g., a large language model), then there should either be a way to access this model for reproducing the results or a way to reproduce the model (e.g., with an open-source dataset or instructions for how to construct the dataset).
            \item We recognize that reproducibility may be tricky in some cases, in which case authors are welcome to describe the particular way they provide for reproducibility. In the case of closed-source models, it may be that access to the model is limited in some way (e.g., to registered users), but it should be possible for other researchers to have some path to reproducing or verifying the results.
        \end{enumerate}
    \end{itemize}

\item {\bf Open access to data and code}
    \item[] Question: Does the paper provide open access to the data and code, with sufficient instructions to faithfully reproduce the main experimental results, as described in supplemental material?
    \item[] Answer: \answerYes{} 
    \item[] Justification: The code is released with instructions at \url{https://github.com/siomvas/fds}.
    \item[] Guidelines:
    \begin{itemize}
        \item The answer \answerNA{} means that paper does not include experiments requiring code.
        \item Please see the NeurIPS code and data submission guidelines (\url{https://neurips.cc/public/guides/CodeSubmissionPolicy}) for more details.
        \item While we encourage the release of code and data, we understand that this might not be possible, so \answerNo{} is an acceptable answer. Papers cannot be rejected simply for not including code, unless this is central to the contribution (e.g., for a new open-source benchmark).
        \item The instructions should contain the exact command and environment needed to run to reproduce the results. See the NeurIPS code and data submission guidelines (\url{https://neurips.cc/public/guides/CodeSubmissionPolicy}) for more details.
        \item The authors should provide instructions on data access and preparation, including how to access the raw data, preprocessed data, intermediate data, and generated data, etc.
        \item The authors should provide scripts to reproduce all experimental results for the new proposed method and baselines. If only a subset of experiments are reproducible, they should state which ones are omitted from the script and why.
        \item At submission time, to preserve anonymity, the authors should release anonymized versions (if applicable).
        \item Providing as much information as possible in supplemental material (appended to the paper) is recommended, but including URLs to data and code is permitted.
    \end{itemize}

\item {\bf Experimental setting/details}
    \item[] Question: Does the paper specify all the training and test details (e.g., data splits, hyperparameters, how they were chosen, type of optimizer) necessary to understand the results?
    \item[] Answer: \answerYes{} 
    \item[] Justification: We do this in Section 3 and Appendix B.
    \item[] Guidelines:
    \begin{itemize}
        \item The answer \answerNA{} means that the paper does not include experiments.
        \item The experimental setting should be presented in the core of the paper to a level of detail that is necessary to appreciate the results and make sense of them.
        \item The full details can be provided either with the code, in appendix, or as supplemental material.
    \end{itemize}

\item {\bf Experiment statistical significance}
    \item[] Question: Does the paper report error bars suitably and correctly defined or other appropriate information about the statistical significance of the experiments?
    \item[] Answer: \answerNo{} 
    \item[] Justification: The reason we do not run multiple seeds is that the computational burden is too high, and cannot be justified. After preliminary runs established the performance variance across seed was much lower compared to the gap between partitioning methods, we were satisfied that the insights from a single seed were correct. This can be attributed to our use of pre-trained models for initialization, which is known to greatly stabilize performance in FL optimization. The comparison of aggregation methods in Section~\ref{sec:algos} reports the mean and standard deviation over three seeds.
    \item[] Guidelines:
    \begin{itemize}
        \item The answer \answerNA{} means that the paper does not include experiments.
        \item The authors should answer \answerYes{} if the results are accompanied by error bars, confidence intervals, or statistical significance tests, at least for the experiments that support the main claims of the paper.
        \item The factors of variability that the error bars are capturing should be clearly stated (for example, train/test split, initialization, random drawing of some parameter, or overall run with given experimental conditions).
        \item The method for calculating the error bars should be explained (closed form formula, call to a library function, bootstrap, etc.)
        \item The assumptions made should be given (e.g., Normally distributed errors).
        \item It should be clear whether the error bar is the standard deviation or the standard error of the mean.
        \item It is OK to report 1-sigma error bars, but one should state it. The authors should preferably report a 2-sigma error bar than state that they have a 96\% CI, if the hypothesis of Normality of errors is not verified.
        \item For asymmetric distributions, the authors should be careful not to show in tables or figures symmetric error bars that would yield results that are out of range (e.g., negative error rates).
        \item If error bars are reported in tables or plots, the authors should explain in the text how they were calculated and reference the corresponding figures or tables in the text.
    \end{itemize}

\item {\bf Experiments compute resources}
    \item[] Question: For each experiment, does the paper provide sufficient information on the computer resources (type of compute workers, memory, time of execution) needed to reproduce the experiments?
    \item[] Answer: \answerYes{} 
    \item[] Justification: Appendix~\ref{app:validation} reports the hardware and time of the semantic labeling pipeline. For the remaining experiments we provide part of this information, since collecting all of it would require estimating it post hoc.
    \item[] Guidelines:
    \begin{itemize}
        \item The answer \answerNA{} means that the paper does not include experiments.
        \item The paper should indicate the type of compute workers CPU or GPU, internal cluster, or cloud provider, including relevant memory and storage.
        \item The paper should provide the amount of compute required for each of the individual experimental runs as well as estimate the total compute. 
        \item The paper should disclose whether the full research project required more compute than the experiments reported in the paper (e.g., preliminary or failed experiments that didn't make it into the paper). 
    \end{itemize}
    
\item {\bf Code of ethics}
    \item[] Question: Does the research conducted in the paper conform, in every respect, with the NeurIPS Code of Ethics \url{https://neurips.cc/public/EthicsGuidelines}?
    \item[] Answer: \answerYes{} 
    \item[] Justification: We have read the NeurIPS Code of Ethics and verified the submission conforms with it.
    \item[] Guidelines:
    \begin{itemize}
        \item The answer \answerNA{} means that the authors have not reviewed the NeurIPS Code of Ethics.
        \item If the authors answer \answerNo, they should explain the special circumstances that require a deviation from the Code of Ethics.
        \item The authors should make sure to preserve anonymity (e.g., if there is a special consideration due to laws or regulations in their jurisdiction).
    \end{itemize}

\item {\bf Broader impacts}
    \item[] Question: Does the paper discuss both potential positive societal impacts and negative societal impacts of the work performed?
    \item[] Answer: \answerYes{} 
    \item[] Justification: We briefly discuss the potential positive impact of better FL algorithm in the conclusion.
    \item[] Guidelines:
    \begin{itemize}
        \item The answer \answerNA{} means that there is no societal impact of the work performed.
        \item If the authors answer \answerNA{} or \answerNo, they should explain why their work has no societal impact or why the paper does not address societal impact.
        \item Examples of negative societal impacts include potential malicious or unintended uses (e.g., disinformation, generating fake profiles, surveillance), fairness considerations (e.g., deployment of technologies that could make decisions that unfairly impact specific groups), privacy considerations, and security considerations.
        \item The conference expects that many papers will be foundational research and not tied to particular applications, let alone deployments. However, if there is a direct path to any negative applications, the authors should point it out. For example, it is legitimate to point out that an improvement in the quality of generative models could be used to generate Deepfakes for disinformation. On the other hand, it is not needed to point out that a generic algorithm for optimizing neural networks could enable people to train models that generate Deepfakes faster.
        \item The authors should consider possible harms that could arise when the technology is being used as intended and functioning correctly, harms that could arise when the technology is being used as intended but gives incorrect results, and harms following from (intentional or unintentional) misuse of the technology.
        \item If there are negative societal impacts, the authors could also discuss possible mitigation strategies (e.g., gated release of models, providing defenses in addition to attacks, mechanisms for monitoring misuse, mechanisms to monitor how a system learns from feedback over time, improving the efficiency and accessibility of ML).
    \end{itemize}
    
\item {\bf Safeguards}
    \item[] Question: Does the paper describe safeguards that have been put in place for responsible release of data or models that have a high risk for misuse (e.g., pre-trained language models, image generators, or scraped datasets)?
    \item[] Answer: \answerNA{} 
    \item[] Justification: This is not applicable to our paper.
    \item[] Guidelines:
    \begin{itemize}
        \item The answer \answerNA{} means that the paper poses no such risks.
        \item Released models that have a high risk for misuse or dual-use should be released with necessary safeguards to allow for controlled use of the model, for example by requiring that users adhere to usage guidelines or restrictions to access the model or implementing safety filters. 
        \item Datasets that have been scraped from the Internet could pose safety risks. The authors should describe how they avoided releasing unsafe images.
        \item We recognize that providing effective safeguards is challenging, and many papers do not require this, but we encourage authors to take this into account and make a best faith effort.
    \end{itemize}

\item {\bf Licenses for existing assets}
    \item[] Question: Are the creators or original owners of assets (e.g., code, data, models), used in the paper, properly credited and are the license and terms of use explicitly mentioned and properly respected?
    \item[] Answer: \answerYes{} 
    \item[] Justification: We have cited the original owners of assets we have used, and have checked that the corresponding licenses allow the assets to be used for research purposes.
    \item[] Guidelines:
    \begin{itemize}
        \item The answer \answerNA{} means that the paper does not use existing assets.
        \item The authors should cite the original paper that produced the code package or dataset.
        \item The authors should state which version of the asset is used and, if possible, include a URL.
        \item The name of the license (e.g., CC-BY 4.0) should be included for each asset.
        \item For scraped data from a particular source (e.g., website), the copyright and terms of service of that source should be provided.
        \item If assets are released, the license, copyright information, and terms of use in the package should be provided. For popular datasets, \url{paperswithcode.com/datasets} has curated licenses for some datasets. Their licensing guide can help determine the license of a dataset.
        \item For existing datasets that are re-packaged, both the original license and the license of the derived asset (if it has changed) should be provided.
        \item If this information is not available online, the authors are encouraged to reach out to the asset's creators.
    \end{itemize}

\item {\bf New assets}
    \item[] Question: Are new assets introduced in the paper well documented and is the documentation provided alongside the assets?
    \item[] Answer: \answerYes{} 
    \item[] Justification: The repository at \url{https://github.com/siomvas/fds} documents the released code and artifacts and states their license, Apache 2.0.
    \item[] Guidelines:
    \begin{itemize}
        \item The answer \answerNA{} means that the paper does not release new assets.
        \item Researchers should communicate the details of the dataset\slash code\slash model as part of their submissions via structured templates. This includes details about training, license, limitations, etc. 
        \item The paper should discuss whether and how consent was obtained from people whose asset is used.
        \item At submission time, remember to anonymize your assets (if applicable). You can either create an anonymized URL or include an anonymized zip file.
    \end{itemize}

\item {\bf Crowdsourcing and research with human subjects}
    \item[] Question: For crowdsourcing experiments and research with human subjects, does the paper include the full text of instructions given to participants and screenshots, if applicable, as well as details about compensation (if any)? 
    \item[] Answer: \answerNA{} 
    \item[] Justification: This is not applicable to our paper.
    \item[] Guidelines:
    \begin{itemize}
        \item The answer \answerNA{} means that the paper does not involve crowdsourcing nor research with human subjects.
        \item Including this information in the supplemental material is fine, but if the main contribution of the paper involves human subjects, then as much detail as possible should be included in the main paper. 
        \item According to the NeurIPS Code of Ethics, workers involved in data collection, curation, or other labor should be paid at least the minimum wage in the country of the data collector. 
    \end{itemize}

\item {\bf Institutional review board (IRB) approvals or equivalent for research with human subjects}
    \item[] Question: Does the paper describe potential risks incurred by study participants, whether such risks were disclosed to the subjects, and whether Institutional Review Board (IRB) approvals (or an equivalent approval/review based on the requirements of your country or institution) were obtained?
    \item[] Answer: \answerNA{} 
    \item[] Justification: This is not applicable to our paper.
    \item[] Guidelines:
    \begin{itemize}
        \item The answer \answerNA{} means that the paper does not involve crowdsourcing nor research with human subjects.
        \item Depending on the country in which research is conducted, IRB approval (or equivalent) may be required for any human subjects research. If you obtained IRB approval, you should clearly state this in the paper. 
        \item We recognize that the procedures for this may vary significantly between institutions and locations, and we expect authors to adhere to the NeurIPS Code of Ethics and the guidelines for their institution. 
        \item For initial submissions, do not include any information that would break anonymity (if applicable), such as the institution conducting the review.
    \end{itemize}

\item {\bf Declaration of LLM usage}
    \item[] Question: Does the paper describe the usage of LLMs if it is an important, original, or non-standard component of the core methods in this research? Note that if the LLM is used only for writing, editing, or formatting purposes and does \emph{not} impact the core methodology, scientific rigor, or originality of the research, declaration is not required.
    \item[] Answer: \answerYes{} 
    \item[] Justification: Use of an LLM is part of our semantic partitioning method. It is documented as a part of the pipeline, we reference the used model, which is open-source, and provide the exact prompts in the appendix.
    \item[] Guidelines:
    \begin{itemize}
        \item The answer \answerNA{} means that the core method development in this research does not involve LLMs as any important, original, or non-standard components.
        \item Please refer to our LLM policy in the NeurIPS handbook for what should or should not be described.
    \end{itemize}

\end{enumerate}

\end{document}